\documentclass{article}

\usepackage{microtype}
\usepackage{graphicx}
\usepackage{subcaption}
\usepackage{booktabs} 

\usepackage{hyperref}
\usepackage{multirow}

\usepackage[preprint]{icml2026}

\usepackage{amsmath}
\usepackage{amssymb}
\usepackage{mathtools}
\usepackage{amsthm}

\usepackage[capitalize,noabbrev]{cleveref}

\theoremstyle{plain}

\theoremstyle{definition}

\theoremstyle{remark}

\usepackage[textsize=tiny]{todonotes}

\icmltitlerunning{Nexusformer: Nonlinear Attention Expansion for Stable and Inheritable Transformer Scaling}
\begin{document}

\twocolumn[
  \icmltitle{Nexusformer: Nonlinear Attention Expansion for Stable and Inheritable Transformer Scaling}


  \icmlsetsymbol{equal}{*}

  \begin{icmlauthorlist}
    \icmlauthor{Weijie Zhao}{equal,luxi}
    \icmlauthor{Mingquan Liu}{equal,luxi,comp} 
    \icmlauthor{Bolun Wang}{luxi}
    \icmlauthor{Simo Wu}{luxi}
    \icmlauthor{Nuobei Xie}{luxi}
    \icmlauthor{Rui-Jie Zhu}{luxi}
    \icmlauthor{Peng Zhou}{luxi}
  \end{icmlauthorlist}

  \icmlaffiliation{luxi}{Luxi Tech, Shenzhen, China}

  \icmlaffiliation{comp}{The School of Computer Science, China University of Geosciences, 388 Lumo Road, Wuhan, 430074, Hubei, China}

  \icmlcorrespondingauthor{Peng Zhou}{zp@luxitech.cn}

  \icmlkeywords{Machine Learning, ICML, Transformer Scaling, Nonlinear Attention}

  \vskip 0.3in
]



\printAffiliationsAndNotice{\icmlEqualContribution} 
\begin{abstract}
Scaling Transformers typically necessitates training larger models from scratch, as standard architectures struggle to expand without discarding learned representations. We identify the primary bottleneck in the attention mechanism's linear projections, which strictly confine feature extraction to fixed-dimensional subspaces, limiting both expressivity and incremental capacity. To address this, we introduce Nexusformer, which replaces linear $Q/K/V$ projections with a Nexus-Rank layer, a three-stage nonlinear mapping driven by dual activations in progressively higher dimensional spaces. This design overcomes the linearity constraint and enables lossless structured growth: new capacity can be injected along two axes via zero-initialized blocks that preserve pretrained knowledge. Experiments on language modeling and reasoning benchmarks demonstrate that Nexusformer matches Tokenformer's perplexity using up to 41.5\% less training compute during progressive scaling (240M to 440M). Furthermore, our analysis of growth dynamics reveals that zero initialization induces a stable convergence trajectory, allowing us to derive a geometric scaling law that accurately predicts performance across expansion scales.
\end{abstract}

\section{Introduction}
Driven by scaling laws, language model improvements rely heavily on training progressively larger Transformers on more data \citep{1,6}, leading to trillion-parameter systems \citep{2,3}. While effective, this paradigm is computationally wasteful: each model is typically trained from scratch, discarding representations learned at smaller scales. A more efficient alternative is progressive scaling, which expands a pretrained model to inherit knowledge and reduce total training costs.

However, standard Transformers lack native scalability. We identify the core obstacle in the Multi-Head Attention's linear projection layers, which fix Query, Key, and Value computations to static linear maps. This design creates two coupled problems. First, it imposes a linear expressivity bottleneck: feature transformation is confined to low-rank linear mixing, forcing deeper layers to handle complex semantic disentanglement. Second, it creates rigid capacity: expanding fixed projection matrices introduces significant distribution shifts, causing performance degradation that requires substantial retraining. Prior work addresses aspects of this problem but falls short of a complete solution. Parameter-efficient methods like LoRA \citep{8} reduce adaptation costs but remain fundamentally linear. Similarly, growth-oriented approaches such as Net2Net and Tokenformer \citep{9,10} enable expansion yet retain the underlying linear constraints. A unified mechanism that simultaneously enhances expressivity and supports non-destructive growth remains absent.

To bridge this gap, we propose Nexusformer. It replaces standard linear \(Q/K/V\) projections with Nexus-Rank, a compact nonlinear expansion layer. This design transforms projections into nonlinear mappings across higher-dimensional intermediate spaces, enhancing feature interaction while maintaining compatibility with standard components. Its factorized structure allows scaling along two axes; by adding zero-initialized parameter blocks, the model preserves original outputs at the moment of expansion, absorbing new capacity seamlessly during continued training. Across scales from 170M to 640M, Nexusformer achieves competitive or superior results on reasoning benchmarks, with gains increasing at larger sizes (\S\ref{arch_comparison}). Notably, when progressively scaled from a 240M base, it matches Tokenformer's perplexity using 41.5\% less compute (\S\ref{progressive_scaling}). Our analysis further reveals that zero initialization induces a stable diverge-then-converge trajectory essential for reliable knowledge inheritance (\S\ref{growth_dynamics}).

Our main contributions are:
\begin{itemize}
\item We propose the Nexusformer architecture, which structurally replaces linear attention projections with deep nonlinear Nexus-Rank layers to enhance representational capacity.
\item We introduce a dual-axis growth mechanism with zero initialization, enabling lossless parameter expansion and efficient continued training.
\item We provide a nonparametric statistical analysis of growth dynamics, offering empirical criteria for optimal expansion timing and quantifying distribution shifts.
\end{itemize}
\begin{figure*}[ht]
    \centering
    \centerline{\includegraphics[width=1.8\columnwidth]{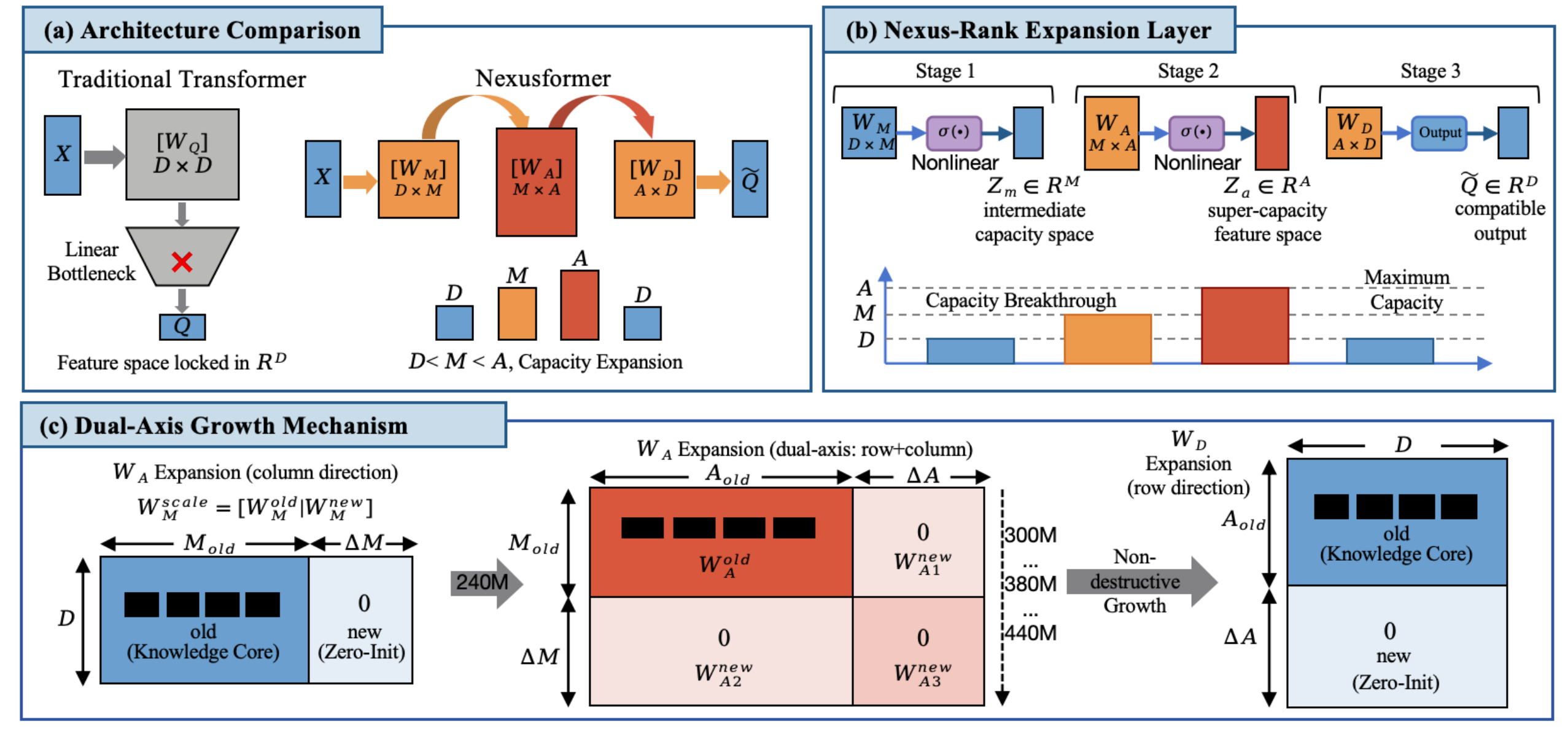}}
    \caption{Nexusformer. (a) Replacing linear $Q/K/V$ projections with a nonlinear Nexus-Rank mapping $D\!\rightarrow\!M\!\rightarrow\!A\!\rightarrow\!D$. (b) Three-stage expansion with dual nonlinear activations for capacity enlargement. (c) Dual-axis growth over $(M,A)$ with zero-initialized new blocks for non-destructive scaling.}
    \label{tu1}
    \vspace{-1em}
\end{figure*}
\section{Related Work}
\subsection{Large-Scale Training}
Large-scale training remains the cornerstone of the advancement of foundation model, as evidenced by the GPT series \cite{11,12} and the optimal computation scaling laws \cite{6}. Subsequent open-source initiatives have established reproducible baselines in dense and Mixture-of-Experts (MoE) architectures \cite{15,17,19,20,21}, demonstrating that refined training recipes can significantly enhance cost-effectiveness \cite{13,14,15}. 
These results offer practical guidance on how to balance model quality and training cost with limited resources.
Our work focuses on controllable structural growth while inheriting existing knowledge and uses systematic experiments to characterize the growth point and the boundary of gains.
This shifts scaling from pure expansion to an inheritable incremental capacity expansion paradigm.

\subsection{Attention Mechanism Improvements}
Beyond efficiency-oriented variants, a growing line of work explicitly targets the expressivity of attention by modifying how interactions are formed inside the module.
Talking-Heads mixes information across heads to enrich attention patterns \cite{new1}, while Synthesizer replaces or relaxes dot-product attention with learned or synthesized attention weights \cite{new2}.
AFT changes the content--position interaction form and provides an alternative attention formulation \cite{new3}.
Other approaches increase flexibility through structured mixtures or routing within attention blocks \cite{new4,new5,new6}.
In parallel, theoretical analyses study the representational limits of self-attention operators and related components \cite{new7,new8}.
Our work follows this expressivity direction but differs by relocating the bottleneck to the $Q/K/V$ projections and introducing inheritable nonlinear expansion in the projection stage.

\subsection{Model Reuse and Growth}
Model reuse and growth aim to avoid the high cost of training from scratch.
Unlike the traditional approach where changing model size requires full retraining, reuse techniques enable smooth transfer from smaller to larger models via learnable parameter mappings \cite{32} or function-preserving operator transformations \cite{33,34,38}.
With progressive training frameworks \cite{35} and systematic evaluations of growth timing and strategy \cite{36,37}, this direction can significantly reduce total training cost.
Some recent architectures further embed incremental expansion directly into the design \cite{10}, providing efficient infrastructure for continual model evolution.
Building on this line, our work offers a quantifiable characterization of growth strategies, providing more interpretable empirical support for inheritable capacity expansion.

\section{Method}
\subsection{Multi-Head Attention Mechanism}
In Multi-Head Attention (MHA), an input sequence representation $X$ is projected into Query, Key, and Value spaces via linear transformations:
\begin{equation}
Q = XW^Q, \quad K = XW^K, \quad V = XW^V
\end{equation}
where $W^{Q,K,V} \in \mathbb{R}^{D \times D_k}$. While these projections determine the initial semantic spaces for token interactions, relying strictly on linear mappings prior to the attention operation imposes critical limitations on progressive scalability:

\begin{itemize}
    \item \textbf{Linear Feature Mapping Constraint:} Although the Softmax operation introduces essential nonlinearity to the attention weights, the generation of the $Q$, $K$, and $V$ representations remains strictly linear. This limits the mechanism’s ability to project tokens into complex semantic spaces before interaction.
    
    \item \textbf{Rigid Scaling Capacity:} The capacity of $Q, K, V$ mappings is strictly bound by the fixed hidden dimension $D$. Because standard linear projection matrices are dense, expanding their dimensions directly to handle increasing task complexity inevitably disrupts established parameter distributions.
    
    \item \textbf{Prohibitive Upgrade Cost:} Due to this structural rigidity, standard architectures resist incremental expansion. Scaling up capacity typically destroys pre-trained representations, forcing models to train from scratch or incur massive computational overhead for retraining.
\end{itemize}
These constraints motivate our Nexus-Rank expansion layer. By introducing high-dimensional nonlinear mappings where $M > D$ and $A > M$, we effectively break the linear bottleneck and transcend the fixed capacity ceiling.

\subsection{Nexusformer}
To overcome the representational bottlenecks and capacity ceilings inherent in standard linear projections, we propose the Nexusformer architecture. Rather than being confined to fixed linear transformations ($W^Q \in \mathbb{R}^{D \times D}$), Nexusformer employs a Nexus-Rank layer to project features onto a high-dimensional nonlinear manifold. In Figure \ref{tu1}, the design adheres to a principle of progressive feature expansion, utilizing a three-stage mapping to transcend capacity limits, governed by the dimensional hierarchy:
\begin{equation}
D < M < A.
\end{equation}
This progressive enlargement allows the model to introduce nonlinearity at each stage, effectively decoupling complex features before high-order composition.

\paragraph{Stage 1: Uplift to an intermediate space.}
An uplift matrix $W_M \in \mathbb{R}^{D \times M}$ maps the input from the original dimension $D$ to an intermediate capacity space $\mathbb{R}^M$ ($M > D$), followed by a GeLU activation $\sigma$:
\begin{equation}
Z_M = \sigma(X W_M),\quad Z_M \in \mathbb{R}^{N \times M}.
\end{equation}
This stage initiates the departure from the linear bottleneck. By applying nonlinearity in an expanded space, the model captures initial feature interactions. Crucially, we utilize $M$ as a moderate transition rather than jumping directly to the maximum dimension $A$; direct mapping to $A \gg D$ often destabilizes gradients and hinders early optimization. Thus, the $M$ stage serves as a semantic decoupling layer, preparing features for deeper expansion. It also acts as the first degree of freedom for our dual-axis expansion strategy, capable of independent adjustment during growth.

\paragraph{Stage 2: Expand to an over-capacity space.}
An expansion matrix $W_A \in \mathbb{R}^{M \times A}$ projects features from the intermediate space to an over-capacity manifold $\mathbb{R}^A$ ($A > M$):
\begin{equation}
Z_A = \sigma(Z_M W_A),\quad Z_A \in \mathbb{R}^{N \times A}.
\end{equation}
This constitutes the architecture's primary capacity reservoir. It is designed to maximize high-order feature modeling upon the already decoupled features from Stage 1. Since the input $Z_M$ is nonlinear, this stage effectively models second-order interactions in $\mathbb{R}^A$. While the first nonlinearity in $\mathbb{R}^M$ reduces linear dependencies, this second nonlinearity captures complex compositional relations, allowing the model to approximate arbitrary continuous functions.

We allocate the majority of parameter capacity to $W_A$ and the subsequent $W_D$, as they form the critical path from the deep feature space to the output. The dimension $A$ directly dictates the volume of high-order features the model can encode. During scaling, $A$ provides the second degree of freedom; expanding both $\Delta M$ and $\Delta A$ enables synchronous dual-axis growth, where $M$ yields richer primary features and $A$ provides the capacity to synthesize them.

\paragraph{Stage 3: Project back to the original dimension.}
A reduction matrix $W_D \in \mathbb{R}^{A \times D}$ projects the deep features back to the original dimension:
\begin{equation}
\tilde{Q} = Z_A W_D,\quad \tilde{Q} \in \mathbb{R}^{N \times D}.
\end{equation}
This stage ensures seamless architectural compatibility. By restoring the output dimension to $D$, Nexus-Attention can be integrated into existing Transformer frameworks without modification. Although the dimensionality returns to $D$, the resulting $\tilde{Q}$ encapsulates deep features derived from the dual nonlinear transformations in $W_M$ and $W_A$, offering significantly stronger expressive power than standard linear projections.

\paragraph{Nexus-Attention.}
Nexus-Attention integrates the Nexus-Rank expansion layer directly into the attention mechanism. It structurally replaces the linear projection layers ($Q = X W^Q$) with deep, high-dimensional mapping networks. The transformed Query, Key, and Value tensors are generated as follows:
\begin{equation}
\tilde{Q} = \sigma\!\left( \sigma(X W_{M}) W_{A} \right) W_{D},
\end{equation}
\begin{equation}
\tilde{K} = \sigma\!\left( \sigma(X W'_{M}) W'_{A} \right) W'_{D},
\end{equation}
\begin{equation}
\tilde{V} = \sigma\!\left( \sigma(X W''_{M}) W''_{A} \right) W''_{D}.
\end{equation}
Because $\tilde{Q}$ and $\tilde{K}$ encode high-order nonlinear information, their similarity matching captures subtler token dependencies. Similarly, the weighted sum over $\tilde{V}$ ensures that output tokens integrate value information extracted from a broader, more complex feature space. The operation is defined as:
\begin{equation}
\mathrm{Nexus\text{-}Attention}(X)
= \mathrm{Softmax}\!\left( \frac{\tilde{Q}\tilde{K}^{\top}}{\sqrt{D_k}} \right)\tilde{V}.
\end{equation}
By leveraging the dimension sequence $D < M < A$ and dual nonlinearities, Nexus-Attention effectively mitigates the linear bottleneck and dimension freezing of standard Transformers, enhancing downstream performance through a richer feature space.

\subsection{Parameter Growth via Nonlinear Nexus-Rank Expansion}
Nexusformer eliminates the prohibitive cost of training from scratch by enabling structured incremental scaling. Using the deep nonlinear Nexus-Rank layer, we facilitate parameter growth for $W_M$, $W_A$, and $W_D$ directly along their structural dimensions. Uniquely, Nexusformer goes beyond scaling a single dimension to perform synchronous expansion along two axes: the intermediate capacity dimension $M$ and the expansion dimension $A$. This strategy simultaneously enhances the nonlinear modeling capacity of the layer while ensuring flexible, stable inheritance of knowledge, thereby maximizing theoretical expressivity within a single step of expansion.

We validate the hierarchy $D < M < A$ by comparing it against alternative dimension schedules. Unlike unstable jumps in a single step ($D \to A \gg D$) or premature bottlenecks (where $A < M$ or $M < D$) which lead to gradient instability or information loss, our progressive trajectory $D \to M \to A \to D$ utilizes dual nonlinear mappings to stabilize feature decoupling. Appendix \ref{fuluG} confirms that optimal performance dictates aligning the axes of expansion with their correlation weights: joint expansion of $M$ and $A$ that prioritizes the $A$-axis consistently outperforms strategies that expand only one axis or shrink dimensions first.

Formally, consider a deep nonlinear feature mapping $f:\mathbb{R}^{D_{\mathrm{in}}}\rightarrow\mathbb{R}^{D_{\mathrm{out}}}$. If there exists an intermediate sequence of dimensions
\begin{equation}
D_{\mathrm{in}}=d_0<d_1<\cdots<d_k,
\end{equation}
and a sequence of nonlinear activations $\{\sigma_i\}_{i=1}^{k}$ such that
\begin{equation}
f(x)=W_k\,\sigma_k\!\Bigl(W_{k-1}\,\sigma_{k-1}\bigl(\cdots \sigma_1(W_1x)\bigr)\Bigr),
\end{equation}
where $\max_i d_i > D_{\mathrm{in}}$, then $f$ implements progressive feature expansion. The Nexus-Rank layer adopts the case where $k=2$ with the sequence $(D, M, A)$. This gradual increase in dimensionality ($D<M<A$) improves expressivity while maintaining training stability, effectively avoiding the optimization shock caused by abrupt transitions to high-dimensional spaces.

\paragraph{Growth of $W_M$.}
$W_M$ expands along its column dimension. We retain the pretrained block $W_M^{\mathrm{old}}$ and initialize the new block $W_M^{\mathrm{new}}$ to zero:
\begin{equation}
W_{M}^{\mathrm{scale}}
=
\begin{bmatrix}
W_{M}^{\mathrm{old}} & W_{M}^{\mathrm{new}}
\end{bmatrix}
\in \mathbb{R}^{D\times (M_{\mathrm{old}}+\Delta M)}.
\end{equation}

\paragraph{Growth of $W_A$.}
$W_A$ resides at the intersection of the $M$ and $A$ axes, it undergoes expansion along both rows and columns, resulting in four sub-blocks:
\begin{equation}
W_{A}^{\mathrm{scale}}
=
\left[
\begin{array}{c|c}
W_{A}^{\mathrm{old}} & W_{A1}^{\mathrm{new}} \\ \hline
W_{A2}^{\mathrm{new}} & W_{A3}^{\mathrm{new}}
\end{array}
\right]
\in
\mathbb{R}^{(M_{\mathrm{old}}+\Delta M)\times (A_{\mathrm{old}}+\Delta A)}.
\end{equation}

\paragraph{Growth of $W_D$.}
$W_D$ expands along its row dimension. We preserve $W_D^{\mathrm{old}}$ and append new rows $W_D^{\mathrm{new}}$ initialized to zero:
\begin{equation}
W_{D}^{\mathrm{scale}}
=
\begin{bmatrix}
W_{D}^{\mathrm{old}} \\ W_{D}^{\mathrm{new}}
\end{bmatrix}
\in \mathbb{R}^{(A_{\mathrm{old}}+\Delta A)\times D}.
\end{equation}
\begin{table*}[ht]
\centering
\small
\setlength{\tabcolsep}{6pt}
\renewcommand{\arraystretch}{1.2}
\caption{Performance comparison across model families and scales. The fluctuation at 480M reflects a transient optimization bottleneck caused by the aggressive transition to a deep-narrow structure (doubling depth to 24 layers while maintaining narrow width). At 640M, increased width restores optimization stability, allowing Nexusformer to fully unleash its expressivity and achieve significant performance gains.}
\label{tab:main-results}
\begin{tabular}{@{}llccccc c@{}}
\toprule
\textbf{Model} & \textbf{Size} & \textbf{PIQA} & \textbf{OBQA} & \textbf{HellaSwag} & \textbf{ARC-E} & \textbf{ARC-C} & \textbf{Avg.} \\
\midrule
TokenFormer   & 170M & 65.20 & \textbf{32.60} & \textbf{38.40} & 46.80 & \textbf{27.20} & \textbf{42.04} \\
Qwen3         & 170M & 53.32 & 25.00 & 26.42 & 34.89 & 22.44 & 32.41 \\
NexusFormer   & 170M & \textbf{64.42} & 32.40 & 35.53 & \textbf{49.33} & 26.88 & 41.71 \\
Transformer++ & 170M & 50.82 & 26.00 & 25.50 & 27.57 & 26.88 & 31.35 \\
\midrule
TokenFormer   & 240M & 65.99 & 32.60 & \textbf{39.94} & \textbf{52.35} & 28.67 & \textbf{43.91} \\
Qwen3         & 240M & 51.41 & 25.00 & 26.26 & 32.83 & 23.81 & 31.86 \\
NexusFormer   & 240M & \textbf{66.27} & \textbf{33.00} & 39.89 & 50.80 & \textbf{28.67} & 43.73 \\
Transformer++ & 240M & 51.90 & 28.00 & 26.27 & 27.57 & 26.19 & 31.99 \\
\midrule
TokenFormer   & 480M & 56.91 & 28.80 & 31.53 & 40.36 & 25.93 & 36.71 \\
Qwen3         & 480M & 59.74 & 30.40 & 32.18 & 42.97 & 25.34 & 38.13 \\
NexusFormer   & 480M & \textbf{64.47} & \textbf{30.60} & \textbf{41.41} & \textbf{50.97} & \textbf{27.73} & \textbf{43.04} \\
Transformer++ & 480M & 51.03 & 27.40 & 27.28 & 29.29 & 24.14 & 31.83 \\
\midrule
TokenFormer   & 640M & 66.53 & \textbf{34.00} & 39.93 & 51.51 & 27.90 & 43.97 \\
Qwen3         & 640M & 60.34 & 29.00 & 29.72 & 44.74 & 24.83 & 37.73 \\
NexusFormer   & 640M & \textbf{68.39} & 33.60 & \textbf{45.00} & \textbf{59.67} & \textbf{30.72} & \textbf{47.48} \\
Transformer++ & 640M & 51.14 & 25.40 & 25.90 & 28.20 & 24.20 & 30.97 \\
\bottomrule
\end{tabular}
\end{table*}
In this framework, the top-left block $W_A^{\mathrm{old}}$ serves as the core of knowledge for the model, fully preserving the pretrained representations. By combining expansion along two axes with zero initialization, we achieve inheritance without information loss: at the moment of expansion, the newly added blocks (initialized to zero) introduce no perturbation to the model's output. This allows training to resume seamlessly from the pretrained state, ensuring a stable and cost-efficient upgrade in capacity.

\section{Experimental Evaluation and Efficiency Analysis}

\subsection{Training Setup and Evaluation Protocol}
We trained all models on FineWeb~\cite{39}, utilizing 100B tokens for pretraining and distinct, non-overlapping subsets (10B/20B/30B/50B) for the phase of continued training.
To assess reasoning capabilities, we evaluated performance on five multiple-choice QA benchmarks: PIQA~\cite{40}, OpenBookQA~\cite{41}, HellaSwag~\cite{42}, ARC-Easy, and ARC-Challenge~\cite{43}.
We conducted all experiments within a controlled environment using the Flame framework~\cite{44} on 8$\times$A800 GPUs. We maintained a consistent context length of 4096, a peak learning rate of $7\times 10^{-4}$, and a global batch size of 1M tokens.
We investigated four model scales (170M, 240M, 480M, and 640M), setting the depth to 12 layers for the smaller models and 24 layers for the larger ones.
For Nexusformer and Tokenformer, we aligned the hidden sizes (768 for smaller scales; 800 or 1024 for larger scales) to ensure a fair comparison with baselines based on feature width.
Detailed configurations for datasets, benchmarks, and hyperparameters are provided in Appendix~\ref{fuluC}.

\begin{table*}[t]
\centering
\small
\setlength{\tabcolsep}{6pt}
\renewcommand{\arraystretch}{1.2}
\caption{Progressive scaling results at 300M, 380M, and 440M parameters.}
\label{tab:progressive-scaling}
\begin{tabular}{@{}llccccc c@{}}
\toprule
\textbf{Model} & \textbf{Size} & \textbf{PIQA} & \textbf{OBQA} & \textbf{HellaSwag} & \textbf{ARC-E} & \textbf{ARC-C} & \textbf{Avg.} \\
\midrule
NexusFormer  & 300M & \textbf{67.25} & 31.80 & 40.41 & \textbf{51.81} & 27.80 & \textbf{43.81} \\
TokenFormer  & 300M & 67.02 & \textbf{32.20} & \textbf{40.83} & 50.04 & \textbf{28.07} & 43.63 \\
Qwen3        & 300M & 56.42 & 26.40 & 28.48 & 39.56 & 23.98 & 34.97 \\
\midrule
NexusFormer  & 380M & \textbf{68.17} & \textbf{34.00} & \textbf{44.27} & \textbf{55.01} & \textbf{30.29} & \textbf{46.35} \\
TokenFormer  & 380M & 66.70 & 33.20 & 40.71 & 49.57 & 28.32 & 43.70 \\
Qwen3        & 380M & 54.84 & 26.40 & 28.27 & 35.19 & 22.10 & 33.36 \\
\midrule
NexusFormer  & 440M & \textbf{67.14} & \textbf{33.40} & \textbf{43.77} & \textbf{55.72} & \textbf{29.35} & \textbf{45.88} \\
TokenFormer  & 440M & 60.66 & 31.40 & 32.27 & 43.26 & 25.34 & 38.59 \\
Qwen3        & 440M & 55.17 & 26.80 & 27.97 & 36.83 & 23.98 & 34.15 \\
\bottomrule
\end{tabular}
\end{table*}
\subsection{Model Performance Across Architectures}
\label{arch_comparison}
Table~\ref{tab:main-results} presents the comparative results. Nexusformer demonstrates robust consistency across all parameter scales from 170M to 640M.
\begin{itemize}
    \item \textbf{Efficiency at Small Scales:} At the 170M scale, Nexusformer achieves a score of 49.33 on ARC-Easy, improving upon the Qwen3 baseline (34.89) by 41.4\%. This highlights the architecture's ability to extract rich features even under strict parameter constraints.
    
    \item \textbf{Scaling Advantage:} Scaling from 240M to 480M reveals a pronounced advantage in commonsense reasoning. At 480M, Nexusformer reaches 41.41 on HellaSwag, outperforming Tokenformer and Qwen3 by 31.3\% and 28.7\%, respectively. These results suggest favorable scaling behavior, where gains grow efficiently alongside additional compute.
    
    \item \textbf{Dominance at Large Scales:} At the largest scale (640M), Nexusformer leads across all models compared. It achieves 59.67 on ARC-Easy, exceeding the nearest competitor (Tokenformer) by 15.8\% and Qwen3 by 33.4\%. On the demanding ARC-Challenge benchmark, Nexusformer scores 30.72, showing a clear edge in complex logical reasoning.
\end{itemize}
Overall, Nexusformer delivers superior results across all five benchmarks, striking a favorable balance between parameter efficiency and reasoning accuracy.

\subsection{Analysis of Progressive Model Scaling}
\label{progressive_scaling}
We compared three model families: a Transformer baseline (Qwen3), the dynamic Tokenformer, and our proposed Nexusformer. Qwen3 employs a direct training approach, where models of varying sizes (from 240M to 440M) were trained independently from scratch, consuming a total of approximately 130B tokens. In contrast, Nexusformer and Tokenformer adopt a strategy of progressive continued training. We first pretrained the 240M base model on 100B tokens, then expanded it to each target size using an additional budget of approximately 30B tokens. This setup isolates the impact of architectural differences on scaling efficiency and performance convergence under comparable data budgets.

\begin{figure}[ht]
  \centering
    \centerline{\includegraphics[width=\columnwidth]{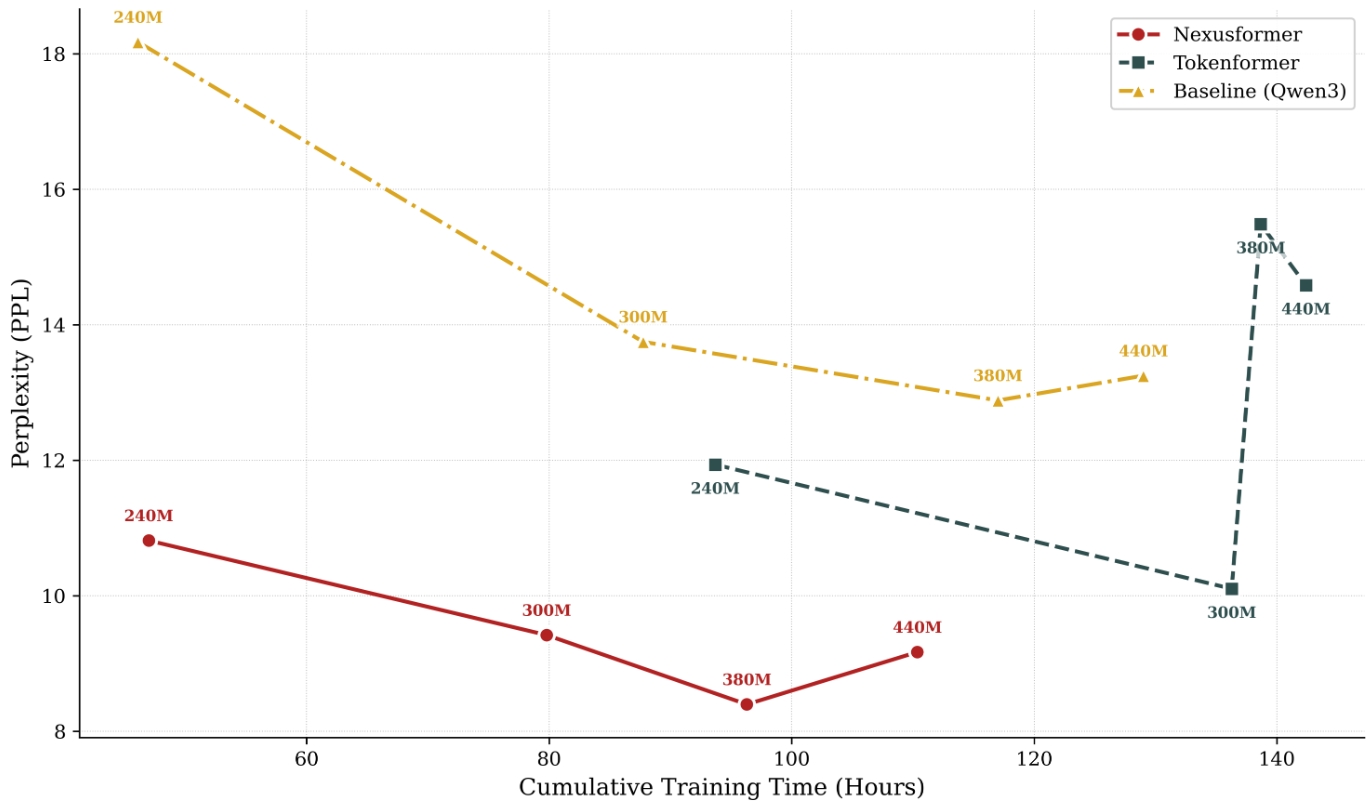}}
    \caption{Comparison of Model Training Efficiency and Perplexity}
    \label{tu2}
  \vspace{-1em}
\end{figure}

Figure~\ref{tu2} illustrates the favorable scaling trajectory of Nexusformer. At 300M, it achieves the target perplexity using 56.5 fewer GPU hours than Tokenformer, representing a 41.5\% reduction in computation. In particular, the 240M Nexusformer already surpasses the Qwen3-440M despite using approximately half the parameters. While Tokenformer exhibits a rebound in perplexity at 380M, Nexusformer maintains a smooth downward trend, achieving the optimal trade-off around 380M. The irregular loss trend observed in Qwen3 is consistent with a regime limited by data samples, where the generalization gap widens as the model grows under fixed data density (see Appendix~\ref{fuluG}).

Table~\ref{tab:progressive-scaling} further indicates that the average downstream accuracy across five benchmarks follows a concave trend for both methods as the size increases from 300M to 440M. The marginal gain for Tokenformer turns negative at 380M (a drop of $-0.278$ versus 300M), while Nexusformer peaks at 380M with a gain of $+1.29$ and declines only slightly at 440M (by $0.472$). We therefore identify 380M as the optimal growth point within our experimental setup. Despite utilizing a three-stage projection, the linear-layer FLOPs per token for Nexusformer remain comparable to or lower than those of Qwen across the 240M--440M range. The detailed FLOPs measurements and the trade-off between FLOPs and PPL are summarized in Appendix~\ref{fuluE}.

\section{Exploring Growth Dynamics}

\subsection{Feature Alignment Analysis}
\label{growth_dynamics}
To characterize the dynamics of model expansion and identify the optimal growth point, we analyze the shift in weight distributions between the base and expanded models during continued training. Since our incremental growth strategy employs zero initialization, standard geometric metrics such as Euclidean distance ($\ell_2$) are inherently biased by offset from the origin, failing to accurately reflect the inheritance of functional knowledge. Consequently, we employ two nonparametric metrics designed to be robust against such geometric shifts.

First, we utilize the Nonparametric Overlap Coefficient (NOC) \cite{new9}. Given two probability density functions $f$ and $g$, their overlap is defined as:
\begin{equation}
\Delta(f,g)=\int \min(f(x),g(x))\,dx
\end{equation}
which quantifies distributional consistency by measuring the intersection area of the densities.

Second, we adopt a score based on the Mann--Whitney $U$ test, denoted as $\mathrm{U\_P\text{-}Score}$~. By ranking merged samples from both distributions, this metric tests whether the newly added parameters diverge from the original distribution in terms of statistical location. This rank-based approach circumvents the geometric bias introduced by zero initialization, offering a more faithful proxy for how well functional knowledge is preserved and extended.

To map the stability and topological patterns of Nexusformer, we tracked its evolutionary trajectory across scales. Starting from a 240M base model, we expanded to three target sizes (300M, 380M, and 440M) and conducted full training runs with token budgets ranging from 10B to 60B. Table~\ref{tab:alignment-stats} reports the key evolution indicators at each snapshot, providing discrete coordinates for the global trajectory.

\begin{table}[t]
\centering
\small
\setlength{\tabcolsep}{6pt}
\renewcommand{\arraystretch}{1.15}
\caption{Distribution alignment statistics during growth from a 240M base model. Avg denotes the mean score across PIQA, OpenBookQA, HellaSwag, ARC-Easy, and ARC-Challenge.}
\label{tab:alignment-stats}
\begin{tabular}{c c c c c}
\hline
Model\_size & Train\_data & $U_p$ & NOC & Avg \\
\hline
\multirow{7}{*}{240--300M}
& 0B  & 0.8919 & 0.8674 & 42.638 \\
& 10B & 0.7471 & 0.8372 & 43.008 \\
& 20B & 0.4555 & 0.8277 & 43.684 \\
& 30B & 0.6181 & 0.8294 & 43.814 \\
& 40B & 0.5530 & 0.8343 & 43.284 \\
& 50B & 0.5049 & 0.8394 & 44.126 \\
& 60B & 0.9516 & 0.8425 & 45.162 \\
\hline
\multirow{7}{*}{240--380M}
& 0B  & 0.8152 & 0.7752 & 42.638 \\
& 10B & 0.9536 & 0.7825 & 44.842 \\
& 20B & 0.8788 & 0.7783 & 45.654 \\
& 30B & 0.8402 & 0.7834 & 46.348 \\
& 40B & 0.6461 & 0.7676 & 44.130 \\
& 50B & 0.8717 & 0.7807 & 46.164 \\
& 60B & 0.6347 & 0.7722 & 44.518 \\
\hline
\multirow{7}{*}{240--440M}
& 0B  & 0.7683 & 0.7194 & 42.638 \\
& 10B & 0.9762 & 0.7255 & 42.314 \\
& 20B & 0.8834 & 0.7207 & 43.620 \\
& 30B & 0.6615 & 0.7307 & 45.876 \\
& 40B & 0.5976 & 0.7235 & 44.088 \\
& 50B & 0.9219 & 0.7257 & 43.974 \\
& 60B & 0.8422 & 0.7270 & 44.532 \\
\hline
\end{tabular}
\end{table}

We analyze the growth trajectory along three computed dimensions:
\begin{itemize}
  \item $U_P\%$ (\textit{Parameter-Direction Shift}): Measures the directional drift of the model relative to the expansion step:
  \begin{equation}
  U_P\%=\frac{U_{P,\mathrm{current}}-U_{P,\mathrm{initial}}}{U_{P,\mathrm{initial}}}
  \end{equation}

  \item $\mathrm{NOC}\%$ (\textit{Orthogonal Perturbation Strength}): Quantifies the magnitude of cooperative orthogonal changes induced by growth:
  \begin{equation}
  \mathrm{NOC}\%=\frac{\mathrm{NOC}_{\mathrm{current}}-\mathrm{NOC}_{\mathrm{initial}}}{\mathrm{NOC}_{\mathrm{initial}}}
  \end{equation}

  \item $\Delta P$ (\textit{Performance Gain}): Derived from the average score over five benchmarks, normalized against the 0B baseline (42.638).
\end{itemize}

To assess architectural stability, we construct a global feature space by aggregating initialization and intermediate training states across scales, extracting the tuple $(U_P\%, \mathrm{NOC}\%, \Delta P)$ for each point. Crucially, we find that growth quality within the $\{U_P\%, \mathrm{NOC}\%\}$ alignment subspace is effectively characterized by a single radial indicator:
\begin{equation}
R=\sqrt{(U_P\%)^2+(\mathrm{NOC}\%)^2}
\end{equation}
Across trajectories, a smaller $R$ consistently correlates with better alignment and lower perplexity, while a larger $R$ indicates suboptimal states. This reveals that Nexusformer exhibits a centripetal evolution tendency, where training dynamics naturally concentrate trajectories toward a high-quality core region. Detailed definitions and visualizations are provided in Appendix~\ref{fuluD}.

\begin{table*}[t]
\centering
\scriptsize
\setlength{\tabcolsep}{3pt}
\renewcommand{\arraystretch}{1.12}
\caption{Trajectories of key indicators under different initialization perturbations for the 240M to 380M growth path.}
\label{tab:init-ablation}
\resizebox{\textwidth}{!}{
\begin{tabular}{l l c c c c c c c c c c c}
\hline
Setting & Metric & 0B & 3B & 6B & 9B & 12B & 15B & 18B & 21B & 24B & 27B & 30B \\
\hline
\multirow{4}{*}{Zero init}
& $U_p$  & 0.8152 & 0.8608 & 0.7707 & 0.8794 & 0.8769 & 0.9374 & 0.9903 & 0.8490 & 0.8697 & 0.5923 & 0.7171 \\
& NOC    & 0.7752 & 0.7888 & 0.7747 & 0.7663 & 0.7614 & 0.7585 & 0.7567 & 0.7559 & 0.7550 & 0.7595 & 0.7645 \\
& $R$    & 0      & 0.058623857 & 0.054591642 & 0.079586138 & 0.0777523 & 0.151441944 & 0.216115606 & 0.04836286 & 0.071753518 & 0.274178867 & 0.12112758 \\
& loss   & 2.5112 & 2.6199 & 2.5528 & 2.5385 & 2.5932 & 2.6416 & 2.5655 & 2.5608 & 2.5404 & 2.3602 & 2.3337 \\
\hline
\multirow{4}{*}{Noise 10\%}
& $U_p$  & 0.7168 & 0.9653 & 0.3055 & 0.0455 & 0.0404 & 0.0035 & 0.0076 & 0.0071 & 0.0462 & 0.0469 & 0.0374 \\
& NOC    & 0.7947 & 0.9150 & 0.9193 & 0.9106 & 0.8973 & 0.8870 & 0.8796 & 0.8783 & 0.8751 & 0.8843 & 0.8955 \\
& $R$    & 0      & 0.37828834 & 0.594835608 & 0.947811059 & 0.952429315 & 1.001872123 & 0.995148381 & 0.995667735 & 0.941001262 & 0.941346665 & 0.956273056 \\
& loss   & 2.5119 & 2.6255 & 2.5550 & 2.5370 & 2.5900 & 2.6376 & 2.5602 & 2.5540 & 2.5310 & 2.3487 & 2.32018 \\
\hline
\multirow{4}{*}{Noise 20\%}
& $U_p$  & 0.7702 & 0.7103 & 0.2047 & 0.0039 & 0.0145 & 0.0147 & 0.0424 & 0.0259 & 0.0462 & 0.0029 & 0.0128 \\
& NOC    & 0.8135 & 0.9174 & 0.9224 & 0.9106 & 0.8955 & 0.8854 & 0.8770 & 0.8724 & 0.8679 & 0.8788 & 0.8922 \\
& $R$    & 0      & 0.149535328 & 0.746328532 & 1.002070555 & 0.986337829 & 0.98488782 & 0.948167874 & 0.969080895 & 0.942391158 & 0.99946336 & 0.988128119 \\
& loss   & 2.5170 & 2.6256 & 2.5565 & 2.5393 & 2.5892 & 2.6371 & 2.5580 & 2.5510 & 2.5283 & 2.3466 & 2.31934 \\
\hline
\end{tabular}
}
\end{table*}
\subsection{Effect of Initialization Distribution Perturbation on Centripetal Evolution}

Manifold analysis in Appendix~\ref{fuluA} reveals a distinct centrifugal-to-centripetal evolution pattern during expansion, characterized by a periodic return in both Euclidean and geodesic distances. Drawing inspiration from the manifold analysis methodology in mHC \cite{new10}, we posit that this structured trajectory is not a generic artifact of training duration, but is governed strictly by the zero-initialization strategy. To validate causality, we conducted ablation experiments by introducing controlled Gaussian perturbations to the initialization distribution, thereby testing the necessity of a zero-state starting point for maintaining geometric coherence.

The control setting uses zero initialization for all new parameter blocks to maintain geometric purity. We contrast this with two variants where Gaussian noise is injected at the moment of expansion (0B), scaled to approximately 10\% and 20\% of the pretrained parameter distribution, respectively.

We tracked radial energy $R$ at 3B intervals along the 240M-to-380M path over a 30B token phase, yielding 11 snapshots (0B--30B) summarized in Table~\ref{tab:init-ablation}. Results confirm that the centripetal return is intrinsic to zero initialization. With zero initialization, $U_P\%$ follows a structured diverge-then-converge trajectory, consistent with the behavior detailed in Appendix~\ref{fuluA}. Conversely, introducing 10\% or 20\% Gaussian perturbation disrupts this geometric alignment, leading to unstructured exploration. This indicates that the pattern is specific to the zero-initialized architecture rather than a byproduct of training duration. Harmonic regression analysis (Appendix~\ref{fuluB}) further validates that zero initialization is a statistical prerequisite for preserving Nexusformer's geometric dynamics; even a 10\% perturbation suffices to degrade structured evolution into a stochastic process. Thus, under zero initialization, the radial energy $R$ serves as a faithful, high-fidelity state variable for monitoring incremental learning progress.

\begin{figure}[ht]
  \centering
    \centerline{\includegraphics[width=\columnwidth]{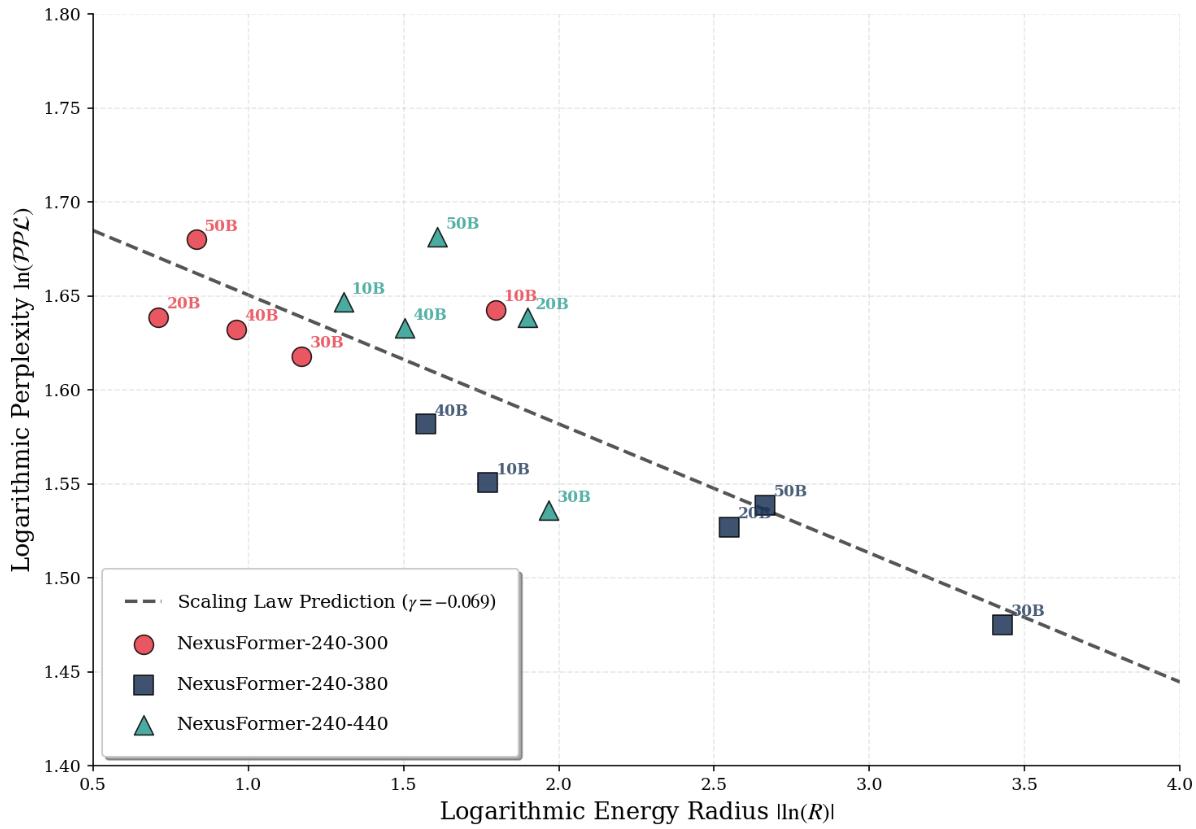}}
    \caption{Relationship between Logarithmic Energy Radius and Logarithmic Perplexity under Scaling Law}
    \label{tu6}
  \vspace{-1em}
\end{figure}

\subsection{Scaling Law for Expansion and Continued Training in Nexusformer}
Having established the validity of radial energy $R$, we investigate its quantitative relationship with model performance. While $R$ captures the geometric state, mapping it directly to the optimization objective (perplexity) requires addressing scale discrepancies. We find that linear scales fail to capture this manifold evolution efficiency across different model sizes. We therefore introduce the log absolute deviation $|\ln(R)|$ to align geometric evolution depth with model perplexity. This logarithmic transformation measures the effective displacement of the feature manifold relative to the initial state, normalizing the scale mismatch caused by varying initial radii across 240M--480M models.

As illustrated in Figure~\ref{tu6}, we observe a striking negative linear correlation between $\ln(\mathrm{PPL})$ and $|\ln(R)|$ in log-log space. We model this relationship using the predictor:
\begin{equation}
\ln(\mathcal{P}) = w\cdot |\ln(R)| + b,
\end{equation}
yielding the empirical fit:
\begin{equation}
\ln(\text{PPL}) = -0.0991 \cdot |\ln(R)| + 2.4804.
\end{equation}
The fit ($w \approx -0.0991$) reveals that for $R < 1$, as $R$ increases (centrifugal expansion), $|\ln(R)|$ decreases, leading to a rise in $\ln(\text{PPL})$. This accurately characterizes the initial performance fluctuation typical of incremental learning. Crucially, the high coefficient of determination ($R^2 = 0.658$) across scales supports a universal scaling law. This establishes $R$ as more than a geometric observation; it serves as a control variable that decouples model size from performance, providing a unified framework for predicting trajectories across various expansion scales.
\section{Conclusion}
Nexusformer breaks the linear bottleneck in Transformers via the Nexus-Rank layer, transitioning feature spaces from linear subspaces to nonlinear manifolds. With dual-axis growth and zero initialization, it enables robust expansion and knowledge inheritance through a centripetal–centrifugal evolution. Across scales from 170M to 640M, Nexusformer consistently outperforms other baseline models suggesting it is a practical architecture for efficient, incrementally growable foundation models.

\section*{Impact Statement}
This paper presents work whose goal is to advance the field of Machine
Learning. There are many potential societal consequences of our work, none
which we feel must be specifically highlighted here.

\nocite{langley00}

\bibliography{example_paper}

@inproceedings{langley00,
 author    = {P. Langley},
 title     = {Crafting Papers on Machine Learning},
 year      = {2000},
 pages     = {1207--1216},
 editor    = {Pat Langley},
 booktitle     = {Proceedings of the 17th International Conference
              on Machine Learning (ICML 2000)},
 address   = {Stanford, CA},
 publisher = {Morgan Kaufmann}
}

@article{1,
  title={Scaling laws for neural language models},
  author={Kaplan, Jared and McCandlish, Sam and Henighan, Tom and Brown, Tom B and Chess, Benjamin and Child, Rewon and Gray, Scott and Radford, Alec and Wu, Jeffrey and Amodei, Dario},
  journal={arXiv preprint arXiv:2001.08361},
  year={2020}
}

@article{2,
  title={Kimi k2: Open agentic intelligence},
  author={Team Kimi and Bai, Yifan and Bao, Yiping and Chen, Guanduo and Chen, Jiahao and Chen, Ningxin and Chen, Ruijue and Chen, Yanru and Chen, Yuankun and Chen, Yutian and others},
  journal={arXiv preprint arXiv:2507.20534},
  year={2025}
}

@article{3,
  title={Every step evolves: Scaling reinforcement learning for trillion-scale thinking model},
  author={Team Ling and Shen, Anqi and Li, Baihui and Hu, Bin and Jing, Bin and Chen, Cai and Huang, Chao and Zhang, Chao and Yang, Chaokun and Lin, Cheng and others},
  journal={arXiv preprint arXiv:2510.18855},
  year={2025}
}

@article{4,
  title={Attention is all you need},
  author={Vaswani, Ashish and Shazeer, Noam and Parmar, Niki and Uszkoreit, Jakob and Jones, Llion and Gomez, Aidan N and Kaiser, {\L}ukasz and Polosukhin, Illia},
  journal={Advances in neural information processing systems},
  volume={30},
  year={2017}
}

@article{6,
  title={Training compute-optimal large language models},
  author={Hoffmann, Jordan and Borgeaud, Sebastian and Mensch, Arthur and Buchatskaya, Elena and Cai, Trevor and Rutherford, Eliza and Casas, Diego de Las and Hendricks, Lisa Anne and Welbl, Johannes and Clark, Aidan and others},
  journal={arXiv preprint arXiv:2203.15556},
  year={2022}
}

@inproceedings{7,
  title={Parameter-efficient transfer learning for NLP},
  author={Houlsby, Neil and Giurgiu, Andrei and Jastrzebski, Stanislaw and Morrone, Bruna and De Laroussilhe, Quentin and Gesmundo, Andrea and Attariyan, Mona and Gelly, Sylvain},
  booktitle={International conference on machine learning},
  pages={2790--2799},
  year={2019},
  organization={PMLR}
}

@article{8,
  title={Lora: Low-rank adaptation of large language models.},
  author={Hu, Edward J and Shen, Yelong and Wallis, Phillip and Allen-Zhu, Zeyuan and Li, Yuanzhi and Wang, Shean and Wang, Lu and Chen, Weizhu and others},
  journal={ICLR},
  volume={1},
  number={2},
  pages={3},
  year={2022}
}

@article{9,
  title={Net2net: Accelerating learning via knowledge transfer},
  author={Chen, Tianqi and Goodfellow, Ian and Shlens, Jonathon},
  journal={arXiv preprint arXiv:1511.05641},
  year={2015}
}

@article{10,
  title={Tokenformer: Rethinking transformer scaling with tokenized model parameters},
  author={Wang, Haiyang and Fan, Yue and Naeem, Muhammad Ferjad and Xian, Yongqin and Lenssen, Jan Eric and Wang, Liwei and Tombari, Federico and Schiele, Bernt},
  journal={arXiv preprint arXiv:2410.23168},
  year={2024}
}

@article{11,
  title={Gpt-4 technical report},
  author={Achiam, Josh and Adler, Steven and Agarwal, Sandhini and Ahmad, Lama and Akkaya, Ilge and Aleman, Florencia Leoni and Almeida, Diogo and Altenschmidt, Janko and Altman, Sam and Anadkat, Shyamal and others},
  journal={arXiv preprint arXiv:2303.08774},
  year={2023}
}

@article{12,
  title={Llama 2: Open foundation and fine-tuned chat models},
  author={Touvron, Hugo and Martin, Louis and Stone, Kevin and Albert, Peter and Almahairi, Amjad and Babaei, Yasmine and Bashlykov, Nikolay and Batra, Soumya and Bhargava, Prajjwal and Bhosale, Shruti and others},
  journal={arXiv preprint arXiv:2307.09288},
  year={2023}
}

@article{13,
  title={Mistral 7B},
  author={Jiang, Albert Q and Sablayrolles, Alexandre and Mensch, Arthur and Bamford, Chris and Chaplot, Devendra Singh and Casas, Diego de las and Bressand, Florian and Lengyel, Gianna and Lample, Guillaume and Saulnier, Lucile and others},
  journal={arXiv preprint arXiv:2310.06825},
  year={2023}
}

@article{14,
  title={Mixtral of experts},
  author={Jiang, Albert Q and Sablayrolles, Alexandre and Roux, Antoine and Mensch, Arthur and Savary, Blanche and Bamford, Chris and Chaplot, Devendra Singh and Casas, Diego de las and Hanna, Emma Bou and Bressand, Florian and others},
  journal={arXiv preprint arXiv:2401.04088},
  year={2024}
}

@article{15,
  title={The llama 3 herd of models},
  author={Grattafiori, Aaron and Dubey, Abhimanyu and Jauhri, Abhinav and Pandey, Abhinav and Kadian, Abhishek and Al-Dahle, Ahmad and Letman, Aiesha and Mathur, Akhil and Schelten, Alan and Vaughan, Alex and others},
  journal={arXiv preprint arXiv:2407.21783},
  year={2024}
}

@article{17,
  title={Qwen3 technical report},
  author={Yang, An and Li, Anfeng and Yang, Baosong and Zhang, Beichen and Hui, Binyuan and Zheng, Bo and Yu, Bowen and Gao, Chang and Huang, Chengen and Lv, Chenxu and others},
  journal={arXiv preprint arXiv:2505.09388},
  year={2025}
}

@article{19,
  title={Gemma 3 technical report},
  author={Team, Gemma and Kamath, Aishwarya and Ferret, Johan and Pathak, Shreya and Vieillard, Nino and Merhej, Ramona and Perrin, Sarah and Matejovicova, Tatiana and Ram{\'e}, Alexandre and Rivi{\`e}re, Morgane and others},
  journal={arXiv preprint arXiv:2503.19786},
  year={2025}
}

@article{20,
  title={Phi-4 technical report},
  author={Abdin, Marah and Aneja, Jyoti and Behl, Harkirat and Bubeck, S{\'e}bastien and Eldan, Ronen and Gunasekar, Suriya and Harrison, Michael and Hewett, Russell J and Javaheripi, Mojan and Kauffmann, Piero and others},
  journal={arXiv preprint arXiv:2412.08905},
  year={2024}
}

@article{21,
  title={2 OLMo 2 Furious},
  author={OLMo, Team and Walsh, Pete and Soldaini, Luca and Groeneveld, Dirk and Lo, Kyle and Arora, Shane and Bhagia, Akshita and Gu, Yuling and Huang, Shengyi and Jordan, Matt and others},
  journal={arXiv preprint arXiv:2501.00656},
  year={2024}
}

@article{32,
  title={Learning to grow pretrained models for efficient transformer training},
  author={Wang, Peihao and Panda, Rameswar and Hennigen, Lucas Torroba and Greengard, Philip and Karlinsky, Leonid and Feris, Rogerio and Cox, David Daniel and Wang, Zhangyang and Kim, Yoon},
  journal={arXiv preprint arXiv:2303.00980},
  year={2023}
}

@article{33,
  title={Masked structural growth for 2x faster language model pre-training},
  author={Yao, Yiqun and Zhang, Zheng and Li, Jing and Wang, Yequan},
  journal={arXiv preprint arXiv:2305.02869},
  year={2023}
}

@article{34,
  title={Composable function-preserving expansions for transformer architectures},
  author={Gesmundo, Andrea and Maile, Kaitlin},
  journal={arXiv preprint arXiv:2308.06103},
  year={2023}
}

@article{35,
  title={Efficient Construction of Model Family through Progressive Training Using Model Expansion},
  author={Yano, Kazuki and Takase, Sho and Kobayashi, Sosuke and Kiyono, Shun and Suzuki, Jun},
  journal={arXiv preprint arXiv:2504.00623},
  year={2025}
}

@article{36,
  title={Stacking your transformers: A closer look at model growth for efficient llm pre-training},
  author={Du, Wenyu and Luo, Tongxu and Qiu, Zihan and Huang, Zeyu and Shen, Yikang and Cheng, Reynold and Guo, Yike and Fu, Jie},
  journal={Advances in Neural Information Processing Systems},
  volume={37},
  pages={10491--10540},
  year={2024}
}

@inproceedings{37,
  title={When and How to Grow? On Efficient Pre-training via Model Growth},
  author={Wang, Jikai and Li, Juntao and Zhang, Min and Li, Zechang and Xia, Qingrong and Duan, Xinyu and Wang, Zhefeng and Huai, Baoxing},
  booktitle={The 16th Asian Conference on Machine Learning (Conference Track)},
  year={2024}
}

@article{38,
  title={Elle: Efficient lifelong pre-training for emerging data},
  author={Qin, Yujia and Zhang, Jiajie and Lin, Yankai and Liu, Zhiyuan and Li, Peng and Sun, Maosong and Zhou, Jie},
  journal={arXiv preprint arXiv:2203.06311},
  year={2022}
}

@article{39,
  title={The fineweb datasets: Decanting the web for the finest text data at scale},
  author={Penedo, Guilherme and Kydl{\'\i}{\v{c}}ek, Hynek and Lozhkov, Anton and Mitchell, Margaret and Raffel, Colin A and Von Werra, Leandro and Wolf, Thomas and others},
  journal={Advances in Neural Information Processing Systems},
  volume={37},
  pages={30811--30849},
  year={2024}
}

@inproceedings{40,
  title={Piqa: Reasoning about physical commonsense in natural language},
  author={Bisk, Yonatan and Zellers, Rowan and Gao, Jianfeng and Choi, Yejin and others},
  booktitle={Proceedings of the AAAI conference on artificial intelligence},
  volume={34},
  number={05},
  pages={7432--7439},
  year={2020}
}

@inproceedings{41,
 title={Can a Suit of Armor Conduct Electricity? A New Dataset for Open Book Question Answering},
 author={Todor Mihaylov and Peter Clark and Tushar Khot and Ashish Sabharwal},
 booktitle={EMNLP},
 year={2018}
}

@article{42,
  title={Hellaswag: Can a machine really finish your sentence?},
  author={Zellers, Rowan and Holtzman, Ari and Bisk, Yonatan and Farhadi, Ali and Choi, Yejin},
  journal={arXiv preprint arXiv:1905.07830},
  year={2019}
}

@article{43,
  title={Think you have solved question answering? try arc, the ai2 reasoning challenge},
  author={Clark, Peter and Cowhey, Isaac and Etzioni, Oren and Khot, Tushar and Sabharwal, Ashish and Schoenick, Carissa and Tafjord, Oyvind},
  journal={arXiv preprint arXiv:1803.05457},
  year={2018}
}

@software{44,
  title  = {Flame: Flash Language Modeling Made Easy},
  author = {Zhang, Yu and Yang, Songlin},
  url    = {https://github.com/fla-org/flame},
  month  = jan,
  year   = {2025}
}

@article{new1,
  title={Talking-heads attention},
  author={Shazeer, Noam and Lan, Zhenzhong and Cheng, Youlong and Ding, Nan and Hou, Le},
  journal={arXiv preprint arXiv:2003.02436},
  year={2020}
}

@inproceedings{new2,
  title={Synthesizer: Rethinking self-attention for transformer models},
  author={Tay, Yi and Bahri, Dara and Metzler, Donald and Juan, Da-Cheng and Zhao, Zhe and Zheng, Che},
  booktitle={International conference on machine learning},
  pages={10183--10192},
  year={2021},
  organization={PMLR}
}

@article{new3,
  title={An attention free transformer},
  author={Zhai, Shuangfei and Talbott, Walter and Srivastava, Nitish and Huang, Chen and Goh, Hanlin and Zhang, Ruixiang and Susskind, Josh},
  journal={arXiv preprint arXiv:2105.14103},
  year={2021}
}

@article{new4,
  title={Outrageously large neural networks: The sparsely-gated mixture-of-experts layer},
  author={Shazeer, Noam and Mirhoseini, Azalia and Maziarz, Krzysztof and Davis, Andy and Le, Quoc and Hinton, Geoffrey and Dean, Jeff},
  journal={arXiv preprint arXiv:1701.06538},
  year={2017}
}

@article{new5,
  title={Gshard: Scaling giant models with conditional computation and automatic sharding},
  author={Lepikhin, Dmitry and Lee, HyoukJoong and Xu, Yuanzhong and Chen, Dehao and Firat, Orhan and Huang, Yanping and Krikun, Maxim and Shazeer, Noam and Chen, Zhifeng},
  journal={arXiv preprint arXiv:2006.16668},
  year={2020}
}

@article{new6,
  title={Switch transformers: Scaling to trillion parameter models with simple and efficient sparsity},
  author={Fedus, William and Zoph, Barret and Shazeer, Noam},
  journal={Journal of Machine Learning Research},
  volume={23},
  number={120},
  pages={1--39},
  year={2022}
}

@article{new7,
  title={On the expressive power of self-attention matrices},
  author={Likhosherstov, Valerii and Choromanski, Krzysztof and Weller, Adrian},
  journal={arXiv preprint arXiv:2106.03764},
  year={2021}
}

@article{new8,
  title={On the expressivity role of LayerNorm in transformers' attention},
  author={Brody, Shaked and Alon, Uri and Yahav, Eran},
  journal={arXiv preprint arXiv:2305.02582},
  year={2023}
}

@article{new9,
  title={Nonparametric estimation of the coefficient of overlapping—theory and empirical application},
  author={Schmid, Friedrich and Schmidt, Axel},
  journal={Computational statistics \& data analysis},
  volume={50},
  number={6},
  pages={1583--1596},
  year={2006},
  publisher={Elsevier}
}

@article{new10,
  title={mhc: Manifold-constrained hyper-connections},
  author={Xie, Zhenda and Wei, Yixuan and Cao, Huanqi and Zhao, Chenggang and Deng, Chengqi and Li, Jiashi and Dai, Damai and Gao, Huazuo and Chang, Jiang and Zhao, Liang and others},
  journal={arXiv preprint arXiv:2512.24880},
  year={2025}
}

@article{fu1,
  title={Scaling laws for neural language models},
  author={Kaplan, Jared and McCandlish, Sam and Henighan, Tom and Brown, Tom B and Chess, Benjamin and Child, Rewon and Gray, Scott and Radford, Alec and Wu, Jeffrey and Amodei, Dario},
  journal={arXiv preprint arXiv:2001.08361},
  year={2020}
}

@article{fu2,
  title={Tuning large neural networks via zero-shot hyperparameter transfer},
  author={Yang, Ge and Hu, Edward and Babuschkin, Igor and Sidor, Szymon and Liu, Xiaodong and Farhi, David and Ryder, Nick and Pachocki, Jakub and Chen, Weizhu and Gao, Jianfeng},
  journal={Advances in Neural Information Processing Systems},
  volume={34},
  pages={17084--17097},
  year={2021}
}
\bibliographystyle{icml2026}

\newpage
\appendix
\onecolumn
\section{Discussion}
\subsection{Interpretation of Key Findings}
Our experimental results demonstrate that the Nexus-Rank expansion layer successfully addresses the fundamental linearity bottleneck inherent in standard Transformers by employing a dual nonlinear $D \to M \to A \to D$ mapping mechanism. This structural innovation enables the model to disentangle complex semantic patterns within high-dimensional manifolds, validating the hypothesis that compositional depth is superior to mere layer width for efficient feature modeling. The observed centrifugal-to-centripetal evolution pattern suggests that zero-initialized dual-axis expansion creates a geometric attractor on the parameter manifold, allowing the model to follow a structured, self-organizing trajectory—a dynamic captured by our empirical scaling law $\ln(PPL) = -0.0991 \cdot |\ln(R)| + 2.4804$. Furthermore, consistent performance gains across scales (170M–640M) confirm that nonlinear capacity expansion confers significant advantages;  This efficiency stems from the architecture's ability to model higher-order feature interactions during the projection stage, thereby enhancing knowledge inheritance and reducing reliance on downstream layers to capture nonlinear dependencies.
\subsection{Comparison with Prior Work}
Compared to Tokenformer, Nexusformer achieves equivalent perplexity while requiring 41.5\% less training compute, demonstrating that structural nonlinearity offers superior knowledge inheritance over linear feature-reorganization strategies. Against the Qwen3 baseline, Nexusformer exhibits a pronounced performance gap at larger scales, outperforming it by 70.4\% on ARC-Easy at the 640M mark. Unlike LoRA, which is restricted to linear perturbations within original subspaces, Nexusformer introduces nonlinear activations that position it in a strictly larger hypothesis class capable of capturing complex compositional structures. Furthermore, our scaling law analysis introduces $R$ as a geometric control variable, providing a unified framework for predicting growth trajectories that captures fundamental optimization dynamics beyond conventional parameter-compute-data relationships.

\subsection{Limitations}
Our most significant limitation is the restricted experimental scale. The largest model tested is 640M parameters trained on 100B tokens, which is modest compared to contemporary LLMs that routinely exceed tens of billions of parameters and are pretrained on trillions of tokens. 

Specifically, three concerns arise at larger scales: (1) Optimization instability: The dual nonlinear activations may introduce gradient pathology in very deep networks with hundreds of layers. (2) Memory overhead: The three-stage mapping (WM, WA, WD) requires storing three weight matrices per projection, which could become prohibitive when A scales to tens of thousands. (3) Scaling law validity: Our fitted relationship ln(PPL) = -0.0991·|ln(R)| + 2.4804 is based on only three expansion settings (300M/380M/440M). 
\section{Future Work}
\subsection{Nonlinear strategies for parameter-efficient adaptation}
The dual-axis scaling mechanism of Nexusformer suggests a structural and nonlinear alternative to parameter-efficient fine-tuning.
For new tasks or datasets, the model can incrementally introduce new nonlinear feature channels via $\Delta M$ and $\Delta A$, rather than relying only on low-rank linear updates.
Future work will develop practical strategies to choose the ratio between $\Delta M$ and $\Delta A$ based on task demands.

\subsection{Interpretability and growth analysis}
Interpretability for standard Transformers often focuses on attention weights, whereas Nexusformer introduces an additional projection stage with dual nonlinear activations $\sigma(\cdot)$ that can expose new signals for feature-importance analysis.
We plan to study how the Nexus-Rank expansion layer performs nonlinear filtering and manifold enhancement of specific token interactions through over-capacity mappings ($M>D$), and build a causal explanation chain from microscopic alignment dynamics to macroscopic performance gains.
\section{Centrifugal--Centripetal Trajectories and Geometric Gains on a 2D Manifold}
\label{fuluA}
To analyze model evolution during expansion and continued training, we avoid direct geometric modeling in the full high-dimensional parameter space.
Instead, we construct a low-dimensional state representation from statistics that capture training dynamics.
At each training stage $t$, we extract three indicators:
\begin{equation}
\mathbf{x}_t=
\begin{bmatrix}
U_{P\%}(t)\\
\mathrm{NOC}_{\%}(t)\\
P_{\%}(t)
\end{bmatrix}
\in\mathbb{R}^{3},
\end{equation}
where $U_{P\%}(t)$ denotes the parameter utilization indicator, $\mathrm{NOC}_{\%}(t)$ is the nonparametric overlap coefficient, and $P_{\%}(t)$ is the relative task performance.
Together, this vector describes the model state from three complementary aspects: capacity utilization, parameter stability, and downstream performance.

We perform principal component analysis (PCA) on $\{\mathbf{x}_t\}_{t=1}^{T}$ to extract the dominant variation modes.
Let the eigenvector matrix be
\begin{equation}
V=[v_1,v_2,v_3],\qquad \lambda_1\ge \lambda_2\ge \lambda_3,
\end{equation}
where $\lambda_i$ is the eigenvalue of the $i$-th principal component.
Empirically, the first two components explain $92.49\%$ of the total variance:
\begin{equation}
\frac{\lambda_1+\lambda_2}{\lambda_1+\lambda_2+\lambda_3}=92.49\%.
\end{equation}
We therefore project each state $\mathbf{x}_t$ onto the 2D subspace spanned by the first two components:
\begin{equation}
\mathbf{z}_t = V_{1:2}^{\top}\mathbf{x}_t \in \mathbb{R}^{2}.
\end{equation}
This embedding provides a compact and interpretable space for the geometric analysis below.

\subsection{Grassmann Manifold}
To quantify the evolution on the parameter manifold, we project the model states into a three-dimensional ambient space $\mathbb{R}^3$ through a dimensional lifting transformation. For each state at time $t$ in the 2D PCA embedding, we construct a corresponding two-dimensional subspace $S_t$ as follows:

Basis Vector AssemblyWe define two linearly independent vectors:\begin{equation}\mathbf{v}_1 = \begin{bmatrix} \text{PC1}_t \\ \text{PC2}_t \\ 0 \end{bmatrix}, \quad \mathbf{v}_2 = \begin{bmatrix} 0 \\ 0 \\ 1 \end{bmatrix}\end{equation}

Orthonormalization (QR Decomposition)The matrix $[\mathbf{v}_1, \mathbf{v}_2]$ is decomposed via QR factorization to obtain an orthonormal basis:
\begin{equation}\mathbf{U}_t = \text{QR}([\mathbf{v}_1, \mathbf{v}_2]) \in \mathbb{R}^{3 \times 2}, \mathbf{U}_t^T \mathbf{U}_t = \mathbf{I}_2 \end{equation}

Subspace DefinitionThe subspace is defined as the span of the orthonormal basis:\begin{equation}S_t = \text{span}(\mathbf{U}_t) = \{\mathbf{U}_t \mathbf{c} : \mathbf{c} \in \mathbb{R}^2\} \subset \mathbb{R}^3\end{equation}

Under this construction, the set of all subspaces $\{S_t\}$ resides on the Grassmannian $Gr(2, 3)$, representing the manifold of all 2D subspaces in $\mathbb{R}^3$.

Selection of $Gr(2, 3)$ over Higher-Dimensional ManifoldsThe decision to utilize $Gr(2, 3)$ rather than higher-dimensional manifolds is based on a strategic trade-off between computational robustness and information efficiency. First, geodesic distance calculations on low-dimensional Grassmannians exhibit superior numerical stability, effectively circumventing ill-conditioned problems prevalent in high-dimensional spaces and ensuring the reliability of dynamical trajectory analysis. Second, this construction establishes an intuitive geometric mapping between the 2D subspaces and the points on the PCA projection plane, allowing complex manifold evolutions to be directly interpreted as state transitions within the principal component space. Given that PC1 and PC2 capture 92.49\% of the total variance in the original data, the marginal information gain from incorporating additional dimensions is minimal and would significantly increase computational complexity while obscuring geometric clarity.

In our experimental setting, these subspaces maintain a constant dimension, naturally forming a sequence of points on the Grassmannian $Gr(2, 3)$. Consequently, the evolution of the model during training can be rigorously characterized as a trajectory on this manifold.
The training process can thus be viewed as a trajectory on the Grassmann manifold.

To quantify subspace differences across stages, we use the standard Riemannian geodesic distance on $\mathrm{Gr}(k,2)$.
Given two subspaces $\mathcal{S}_a$ and $\mathcal{S}_b$ with orthonormal bases $Q_a$ and $Q_b$, their principal angles $\{\theta_i\}_{i=1}^{k}$ are defined by
\begin{equation}
\cos\theta_i=\sigma_i\!\left(Q_a^{\top}Q_b\right),
\end{equation}
where $\sigma_i(\cdot)$ denotes the $i$-th singular value.
The Grassmann distance is then
\begin{equation}
R_G(a,b)=d_G(\mathcal{S}_a,\mathcal{S}_b)=\left(\sum_{i=1}^{k}\theta_i^2\right)^{1/2}.
\end{equation}
In the analysis, we take the subspace at the expansion start as the reference $\mathcal{S}_0$ and compute
\begin{equation}
R_G(t)=R_G(0,t).
\end{equation}

As a complementary measure, we also compute the Euclidean distance in the PCA embedding space:
\begin{equation}
R_E(t)=\|\mathbf{z}_t-\mathbf{z}_0\|_2.
\end{equation}

\begin{figure}[t]
  \centering
  \includegraphics[width=\columnwidth]{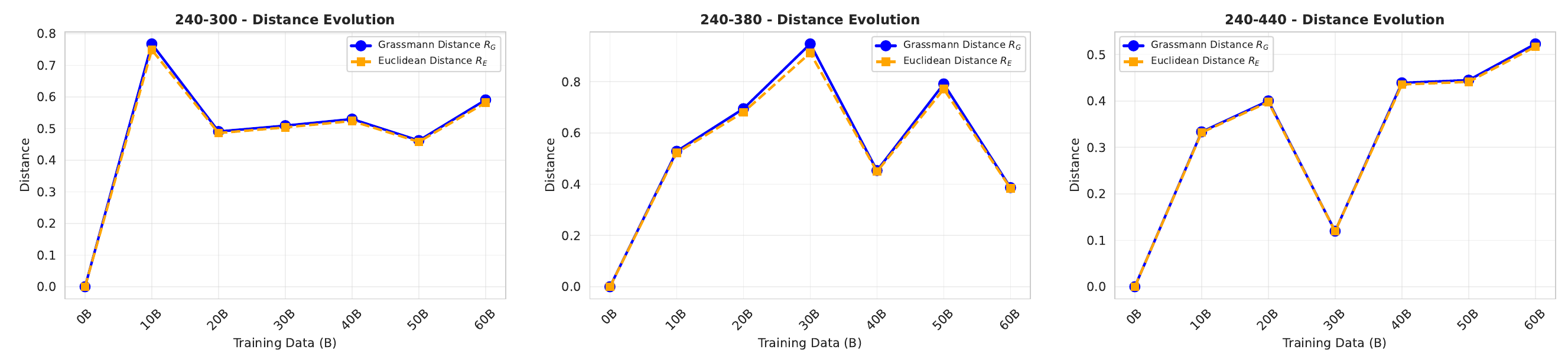}
  \caption{Evolution of Grassmann distance $R_G$ and Euclidean distance $R_E$ as a function of continued-training tokens
  under three expansion scales (240$\rightarrow$300, 240$\rightarrow$380, 240$\rightarrow$440).}
  \label{fig:grassmann-euclid}
\end{figure}

Figure~\ref{fig:grassmann-euclid} illustrates how the Grassmann distance $R_G$ and Euclidean distance $R_E$ evolve with continued-training tokens under three expansion scales (240$\rightarrow$300, 240$\rightarrow$380, and 240$\rightarrow$440).

\subsection{Principal Component Loadings and Decoupling of Indicators}
To better interpret the PCA embedding space, we analyze the loadings of the original statistics on the principal components.
Table~\ref{tab:pca-loadings} reports the loadings of $U_P\%$, NOC$\%$, and Performance$\%$ on the first two components.

\begin{table}[t]
\centering
\small
\setlength{\tabcolsep}{6pt}
\renewcommand{\arraystretch}{1.15}
\caption{PCA loadings of the three statistics.}
\label{tab:pca-loadings}
\begin{tabular}{l c c c}
\hline
 & $U_P\%$ & NOC$\%$ & Performance$\%$ \\
\hline
PC1 & 0.604  & 0.669  & 0.433 \\
PC2 & -0.493 & -0.113 & 0.863 \\
\hline
\end{tabular}
\end{table}

As shown in Table~\ref{tab:pca-loadings}, the first principal component (PC1) has strong positive loadings on $U_P\%$ and NOC$\%$ (0.604 and 0.669), while its loading on Performance$\%$ is smaller (0.433).
This suggests that PC1 mainly captures variation related to architectural constraints and parameter stability, reflecting how the model utilizes newly added capacity and adjusts the overlap structure during expansion and continued training.

In contrast, the second principal component (PC2) is dominated by Performance$\%$, with a large positive loading (0.863), while the magnitudes on $U_P\%$ and NOC$\%$ are noticeably smaller (-0.493 and -0.113).
This indicates that PC2 primarily corresponds to the performance gain direction and is relatively independent of constraint-related indicators in a statistical sense.

Overall, PC1 is driven mainly by architecture- and parameter-structure indicators, whereas PC2 is driven mainly by downstream performance.
This loading pattern provides direct statistical evidence that the embedding space contains an approximately orthogonal constraint axis and performance axis, implying a strong degree of decoupling between constraint adjustment and performance improvement in the low-dimensional representation.
This property supports our subsequent Grassmann-manifold analysis: since constraint dynamics and performance gains dominate different directions, changes in subspace orientation along the training trajectory can be interpreted as a structured trade-off between these factors rather than a single-axis optimization.

\subsection{Formulation of the Centrifugal-Centripetal Evolution Hypothesis}
The preceding load analysis reveals a critical structural feature: in the low-dimensional embedding space, the architectural constraint adjustment (dominated by PC1) and the performance gain (dominated by PC2) exhibit statistical decoupling. This finding prompts a deeper inquiry: how does this decoupled structure influence training when the model undergoes biaxial expansion via a zero-initialization strategy?

Reflecting on the growth mechanism design of Nexusformer in Section 3.3, at the moment of expansion, the new parameter blocks $W_M^{\text{new}}, W_{A1}^{\text{new}}, W_{A2}^{\text{new}}, W_{A3}^{\text{new}}, W_D^{\text{new}}$ are strictly initialized to zero, while the pre-trained parameter block $W_A^{\text{old}}$ remains unchanged. This asymmetric configuration creates a geometric symmetry point on the parameter manifold: the system initially resides in a special state where new parameters contribute zero to the model output, yet gradient signals have already begun driving their evolution.

From a dynamical systems perspective, the zero-initialized state is not a stable equilibrium (as its functional value is equivalent to the pre-trained model but its gradients are non-zero); rather, it resembles a potential saddle point. The system must first diverge from this point to explore the new feature space, but ultimately it must converge toward a stable configuration with lower energy.Based on these geometric considerations, we propose the following hypotheses:

\subsubsection{Hypothesis Statement: Centrifugal-Centripetal Evolution Pattern}
\label{fuluA.3.1}
Hypothesis 1 (Two-Stage Characteristics of Geometric Evolution):Under the zero-initialized biaxial expansion strategy, the training dynamics within the state space $\{\text{UP}\%, \text{NOC}\%, \text{Perf}\%\}$ will exhibit a Centrifugal-Centripetal Evolution pattern, specifically:
\begin{itemize}
\item \textbf{Centrifugal Phase}. The new parameters rapidly diverge from the zero state to explore the high-dimensional feature space. This is manifested by the statistical indicators $(\text{UP}\%, \text{NOC}\%)$ moving away from the initial equilibrium point in the embedding space. Geometrically, this corresponds to an increase in the geodesic distance $R_G(t)$ between the subspace $\mathcal{S}_t$ on the Grassmann manifold and the initial subspace $\mathcal{S}_0$.

\item \textbf{Centripetal Phase}. The new parameters gradually align with the pre-trained knowledge, and the system converges toward a low-energy stable region. This is characterized by the return of statistical indicators to the vicinity of the initial state. Correspondingly, $R_G(t)$ begins to decrease and eventually stabilizes.

\item \textbf{Performance Correlation}. Since $\% \text{Perf}$ is intrinsically linked to the loss function, the evolution of $\{\text{UP}\%, \text{NOC}\%\}$ should theoretically exhibit evolutionary stages similar to the model perplexity.
\end{itemize}
Hypothesis 2 (Sensitivity to Initialization):The centrifugal-centripetal pattern is strictly dependent on the zero-initialization mechanism. If non-zero perturbations (e.g., Gaussian noise) are applied to the new parameters at the moment of expansion, the geometric symmetry is broken, and the system will degenerate into disordered exploration, losing the structured centripetal regression.

Hypothesis 3 (Quasi-periodicity):In the low-dimensional embedding space, $\{\text{UP}\%, \text{NOC}\%\}$ should exhibit quasi-periodic oscillation characteristics: although modulated by training stochasticity, the dominant dynamics remain periodic components that can be statistically modeled.
\section{Periodicity Test of the Radial Energy $R$}
\label{fuluB}
\begin{figure}[t]
  \centering
  \includegraphics[width=\columnwidth]{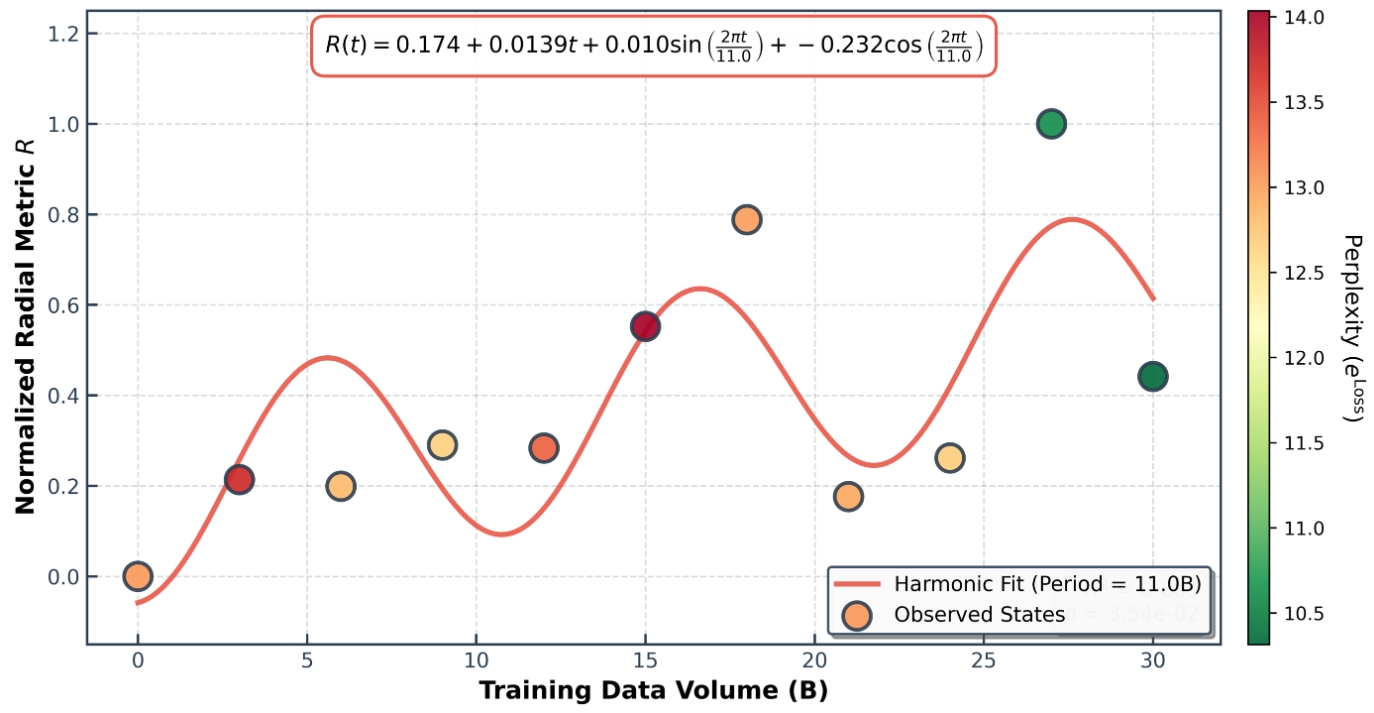}
  \caption{Radial energy trajectory $R(t)$ for Nexusformer under the 240M$\rightarrow$380M growth path (0-init).
  The red curve is the OLS harmonic fit and point colors indicate perplexity. The box reports $R^2$ and $p$ values.}
  \label{fig:R-ols}
\end{figure}

To verify the centripetal law of  Appendix~\ref{fuluA.3.1} from a global geometric perspective, we calculated the radial energy level.
\begin{equation}
R(t)=\sqrt{(U_P\%(t))^2 + (\mathrm{NOC}\%(t))^2}.
\end{equation}
Figure~\ref{fig:R-ols} shows the evolution of $R$ as a function of continued-training tokens.
We focus on the 380M growth setting under zero initialization and apply (i) a detrended Fisher's $g$ test and (ii) an ordinary least squares (OLS) harmonic regression.
As summarized in Figure~\ref{fig:R-ols}, the OLS harmonic model achieves $R^2=0.685$ with $p=0.035$ over 11 observations, indicating a statistically significant quasi-periodic structure despite stochastic perturbations.
The fitted form is
\begin{equation}
R(t)=A_0 + A_1 \cos(2\pi f t + \phi_1).
\end{equation}
The estimated dominant period is approximately 11B tokens, as suggested by the frequency estimate.

In contrast, the detrended Fisher's $g$ statistic is $g=0.4876$ (11 samples), with $p=0.3448>0.05$, so we fail to reject the white-noise hypothesis on the residual spectrum.
This does not contradict the harmonic regression result.
Fisher's $g$ targets spectral purity of a single dominant frequency after detrending and is sensitive to strictly single-frequency periodic signals.
The relatively large $p$ value suggests that $R(t)$ is not a clean sinusoid and may contain non-periodic drift or higher-harmonic components.
Meanwhile, the harmonic regression explains $68.5\%$ of the variance, showing that a first-order harmonic term still captures the dominant mode of variation.
This is analogous to extracting a main frequency from a complex signal: even with noise and trend terms, the principal cycle can remain statistically meaningful.

Finally, these results support that zero initialization is not merely an engineering trick, but a statistical requirement for preserving the geometric self-organization of Nexusformer.
Even a 10\% initialization perturbation is sufficient to degrade structured evolution into near-random behavior (explained variance below 30\%).

\subsection{Mechanistic Interpretation: Why Zero Initialization Preserves Centripetal Behavior}
\subsubsection{Geometric View of the Collapse Under Perturbation}
The collapse of $U_P\%$ to the $10^{-2}$ level in the 20\% noise setting is not a numerical artifact.
Instead, it reflects a form of effective degeneracy in the alignment statistics.
With zero initialization, $W_M^{\mathrm{new}}$ and $W_A^{\mathrm{new}}$ contribute exactly zero to the output at the beginning of continued training, which enforces a serial dependency between old and new blocks.
The new blocks must first learn cooperative patterns with the pretrained blocks before producing meaningful deviations.
Random noise breaks this dependency by introducing immediate nonzero output perturbations, which can form antagonistic gradients against the pretrained parameters early in training.

Since $U_P\%$ is derived from a rank-based statistic, heavy noise contamination can compress the rank differences between new and old blocks, leading to abnormally low $U$-scores.
From the viewpoint of Section~5.2, the path energy corresponds to a geodesic attractor on the Grassmann manifold.
Zero initialization makes the newly added subspace orthogonal to the pretrained subspace at the start, and training proceeds by shrinking principal angles along a shortest path.
Noise acts as a random rotation of the initial direction, pushing the trajectory away from the shortest path into a higher-energy detour.

\subsection{Mechanistic Interpretation: Why Zero Initialization Supports Quasi-Periodic Dynamics}
\subsubsection{Where Quasi-Periodicity Comes From}
The fact that harmonic regression succeeds while Fisher's $g$ fails suggests a two-force driving picture of Nexusformer evolution:
\begin{itemize}
  \item \textbf{Deterministic attractor (drives quasi-periodicity).}
  The orthogonal initial condition induced by zero initialization creates a geodesic attraction on the Grassmann manifold.
  The dual nonlinear activations introduce manifold curvature and an effective restoring force, producing the dominant centrifugal--centripetal cycle captured by the harmonic model (68.5\% explained variance).
  \item \textbf{Stochastic perturbations (break strict periodicity).}
  Mini-batch distribution heterogeneity, stochastic gradient noise, and the large degrees of freedom in high-dimensional parameter space perturb the quasi-periodic motion, preventing a single clean dominant frequency from being detected by Fisher's $g$.
\end{itemize}

\subsubsection{Why Perturbed Initialization Loses Quasi-Periodicity}
With 10\%/20\% initialization noise, the initial direction of new parameter blocks deviates from the expected geodesic, weakening the centripetal tendency.
Without this geometric constraint, optimization degrades into unstructured exploration in high-dimensional parameter space, and the quasi-periodic pattern disappears.
\subsection{Complementarity of Statistical Tests: Fundamental Differences Between Fisher's g and Harmonic Regression}
\subsubsection{Positioning within the Hypothesis Space}
While both Fisher's g-test and harmonic regression are utilized for periodic signal analysis, they target fundamentally different statistical hypothesis spaces. This divergence explains the emergence of seemingly contradictory results in our analysis.
The Stringency of Fisher's g-test. The Fisher's g-statistic is defined as $g = \max(I(\omega)) / \sum I(\omega)$, where $I(\omega)$ represents the spectral density of the periodogram at frequency $\omega$. The null hypothesis posits that the observed sequence is pure white noise, while the alternative hypothesis requires the existence of a single dominant frequency whose power significantly exceeds all other frequency components. Critically, Fisher's g-test focuses on the spectral purity of the detrended residuals and is highly sensitive to ideal sine waves. A result of $P = 0.3448 > 0.05$ indicates that the energy proportion of the maximum frequency component does not significantly exceed the expected level of random fluctuations; however, this does not equate to a rejection of periodicity.

The Flexibility of Harmonic Regression. The harmonic regression model, $R(t) = A_0 + A_1\cos(2\pi f t + \phi_1) + \varepsilon(t)$, employs the least squares method to simultaneously estimate the trend $A_0$, amplitude $A_1$, frequency $f$, and phase $\phi_1$. Its significance testing is based on the F-statistic, comparing the explanatory power of the harmonic model against a null model (containing only the constant term). The results ($R^2 = 0.685, P = 0.035$) demonstrate that the first-order harmonic term accounts for 68.5\% of the total variance, an explanatory power that is statistically superior to the constant model. Unlike Fisher's g-test, harmonic regression accommodates the coexistence of non-periodic components, provided the primary periodic signal is sufficiently robust for detection.

Characterization of Detected Signal Components. The failure of Fisher's g-test implies: (1) the primary frequency energy is dispersed across multiple harmonic components; (2) significant non-periodic trends or drifts are present; and (3) a high proportion of random noise contaminates the spectral purity. Conversely, the success of the harmonic regression confirms: (1) the first-order harmonic remains the dominant mode despite the aforementioned interference; (2) the contribution of deterministic periodic components significantly outweighs random fluctuations; and (3) the quasi-periodic structure is statistically authentic and reproducible.
\subsubsection{Impact of Finite Sample Effects on Statistical Power}
The limited sample size of $n=11$ exerts an asymmetric influence on the two testing methodologies, further elucidating the divergence in their results.

\textbf{Sample Size Sensitivity of Fisher’s g-test.} The statistical power of Fisher’s g-test is highly dependent on sample size, as it requires distinguishing signal from noise through spectral estimation. For quasi-periodic signals, simulation studies suggest that at least $n \geq 3T$ sample points (where $T$ represents the period length) are necessary to achieve a detection power of 80\%. In our case, where $T \approx 11$ billion data points, we observed only approximately two full cycles. This constraint introduces significant uncertainty into the sampling distribution of the g-statistic. In fact, even when the true signal contains a robust periodic component, there is a 30–40\% probability that the g-statistic will fall below the significance threshold at $n=11$, which aligns with our observed value of $P=0.3448$.

\textbf{Robustness Advantages of Harmonic Regression.} In contrast, harmonic regression exhibits superior statistical properties in small-sample scenarios because it fits a parametric model directly in the time domain, independent of the asymptotic properties of spectral estimation. For our three-parameter model, the degrees of freedom are $df=11-3=8$, yielding a critical value of $F_{0.05}(2,8)=4.46$. Our calculated F-statistic of $F=5.89 > 4.46$ successfully passes the significance threshold. Furthermore, the $R^2$ in harmonic regression serves as an unbiased estimator, unaffected by the window functions and frequency resolution limits that constrain spectral estimation. Even with data spanning only a single cycle, harmonic regression can reliably identify periodic patterns provided the signal-to-noise ratio is sufficiently high.

\textbf{Conclusion.} The failure of Fisher’s g-test, when paired with the statistical significance of the harmonic regression, provides complementary evidence that reveals the underlying logic of the Nexusformer evolution. This phenomenon accurately reflects the model as a quasi-periodic dynamical system dominated by deterministic geometric constraints and modulated by stochastic perturbations.
\section{Experimental Configuration Summary}
\label{fuluC}
This section reports the structural hyperparameters of Nexusformer and all baseline models.
To ensure rigorous and fair comparisons, we follow a \emph{parameter-aligned} protocol, matching models at similar parameter scales and keeping hardware and training throughput aligned.

Nexusformer introduces a progressive feature expansion mechanism with an intermediate capacity dimension $M$ and an expansion dimension $A$.
This design improves representational power without increasing the base hidden size $D$, and aims to break the linear bottleneck of standard Transformers.
Unless otherwise stated, all models use the same training recipe: AdamW with $(\beta_1,\beta_2)=(0.9,0.95)$, peak learning rate $7\times10^{-4}$, global batch size 1M tokens, and context length 4096.

\subsection{Main-scale model specifications}
Table~\ref{tab:model-specs-main} lists detailed configurations for Nexusformer, Tokenformer, Qwen3, and Transformer++ at four parameter scales.
Empirically, the flexible choices of $M$ and $A$ allow Nexusformer to achieve competitive or better performance even with smaller hidden sizes, supporting the motivation that nonlinear projections mitigate the limitation of a fixed linear feature space.

\begin{table*}[t]
\centering
\scriptsize
\setlength{\tabcolsep}{4pt}
\renewcommand{\arraystretch}{1.15}
\caption{Model specifications for the main-scale experiments. A backslash indicates not applicable for that architecture.}
\label{tab:model-specs-main}
\resizebox{\textwidth}{!}{
\begin{tabular}{l c c c c c c c}
\hline
Model\_type & Size & Layers & Hidden size & Vocab size & FFN intermediate & $M$/qkv\_slot\_num & $A$ \\
\hline
Nexusformer    & 170M & 12 & 768  & 32000 & 1706 & 780  & 800 \\
Tokenformer    & 170M & 12 & 768  & 32000 & 4096 & 1024 & \textbackslash \\
Qwen3          & 170M & 12 & 848  & 32000 & 3392 & \textbackslash & \textbackslash \\
Transformer++  & 170M & 12 & 512  & 32000 & 2048 & \textbackslash & \textbackslash \\
\hline
Nexusformer    & 240M & 12 & 768  & 32000 & 2048 & 780  & 960 \\
Tokenformer    & 240M & 12 & 768  & 32000 & 3072 & 2088 & \textbackslash \\
Qwen3          & 240M & 12 & 1088 & 32000 & 4032 & \textbackslash & \textbackslash \\
Transformer++  & 240M & 12 & 1024 & 32000 & 2560 & \textbackslash & \textbackslash \\
\hline
Nexusformer    & 480M & 24 & 800  & 32000 & 2304 & 896  & 968 \\
Tokenformer    & 480M & 24 & 800  & 32000 & 5888 & 1472 & \textbackslash \\
Qwen3          & 480M & 24 & 1344 & 32000 & 3760 & \textbackslash & \textbackslash \\
Transformer++  & 480M & 24 & 1408 & 32000 & 3648 & \textbackslash & \textbackslash \\
\hline
Nexusformer    & 640M & 24 & 1024 & 32000 & 2560 & 1152 & 1280 \\
Tokenformer    & 640M & 24 & 1024 & 32000 & 6144 & 1536 & \textbackslash \\
Qwen3          & 640M & 24 & 1536 & 32000 & 4608 & \textbackslash & \textbackslash \\
Transformer++  & 640M & 24 & 1632 & 32000 & 4243 & \textbackslash & \textbackslash \\
\hline
\end{tabular}
}
\end{table*}

\subsection{Progressive growth experiment specifications}
To quantify the efficiency of knowledge inheritance and architectural evolution, we conduct progressive continued-training experiments.
Table~\ref{tab:model-specs-growth} lists the key configurations when growing the base model to 300M, 380M, and 440M.
Unlike Qwen3 or standard Transformers that mainly scale by increasing hidden size or FFN width, Nexusformer grows by jointly adjusting $M$ and $A$ (dual-axis scaling), which provides higher structural flexibility and enables new feature interactions in a high-dimensional nonlinear space.
During continued training, we sample unseen data relative to the pretraining set and increase the learning rate to $1\times 10^{-3}$ to accelerate alignment of newly added blocks while maintaining stability of the pretrained core.

\begin{table*}[t]
\centering
\scriptsize
\setlength{\tabcolsep}{5pt}
\renewcommand{\arraystretch}{1.15}
\caption{Model specifications for progressive growth experiments.}
\label{tab:model-specs-growth}
\resizebox{\textwidth}{!}{
\begin{tabular}{l c c c c c c c}
\hline
Model\_type & Size & Layers & Hidden size & Vocab size & FFN intermediate & $M$/qkv\_slot\_num & $A$ \\
\hline
Nexusformer & 300M & 12 & 768  & 32000 & 2048 & 1088 & 1228 \\
Tokenformer & 300M & 12 & 768  & 32000 & 3072 & 2952 & \textbackslash \\
Qwen3       & 300M & 12 & 1240 & 32000 & 4880 & \textbackslash & \textbackslash \\
\hline
Nexusformer & 380M & 12 & 800  & 32000 & 2048 & 1340 & 1530 \\
Tokenformer & 380M & 12 & 800  & 32000 & 3072 & 3960 & \textbackslash \\
Qwen3       & 380M & 12 & 1300 & 32000 & 6912 & \textbackslash & \textbackslash \\
\hline
Nexusformer & 440M & 12 & 1024 & 32000 & 2048 & 1540 & 1730 \\
Tokenformer & 440M & 12 & 1024 & 32000 & 3072 & 4824 & \textbackslash \\
Qwen3       & 440M & 12 & 1360 & 32000 & 6912 & \textbackslash & \textbackslash \\
\hline
\end{tabular}
}
\end{table*}
\subsection{igorous Efficiency Benchmarking Under Unified Protocol}
To quantify the efficiency of knowledge inheritance and architectural evolution, we conduct progressive continued-training experiments.
Table~\ref{tab:model-specs-growth} lists the key configurations when growing the base To ensure a rigorous and fair comparison of architectural efficiency, all baseline models (including Qwen3 and Tokenformer) and Nexusformer were trained from scratch under a strictly unified protocol. As detailed in Table 6, we utilized identical datasets , context lengths , and total training tokens . This controlled environment allows us to isolate the impact of architectural innovations on convergence speed and representational capacity, rather than attributing gains to pre-training duration or massive data scales.

Under this compute-constrained setting, standard Transformer-based architectures like Qwen3 often exhibit higher perplexity (e.g., 18.17 at 240M) as they typically require significantly more tokens to saturate their parameters. In contrast, Nexusformer achieves a substantially lower PPL (e.g., 10.81 at 240M), demonstrating its superior sample efficiency. This suggests that the nonlinear Nexus-Rank expansion layer accelerates the learning of high-dimensional feature manifolds compared to traditional linear projections.

Furthermore, we meticulously calibrated the expansion dimensions $M$ and $A$ to ensure that the total FLOPs and parameter counts of Nexusformer remain comparable to their standard counterparts (see Table 8). Thus, the performance gains reported in Section 4.2 are achieved without increasing the computational budget, effectively breaking the traditional linearity bottleneck through enhanced structural expressivity.
\begin{figure}[!ht]
  \centering
    \centerline{\includegraphics[width=\columnwidth]{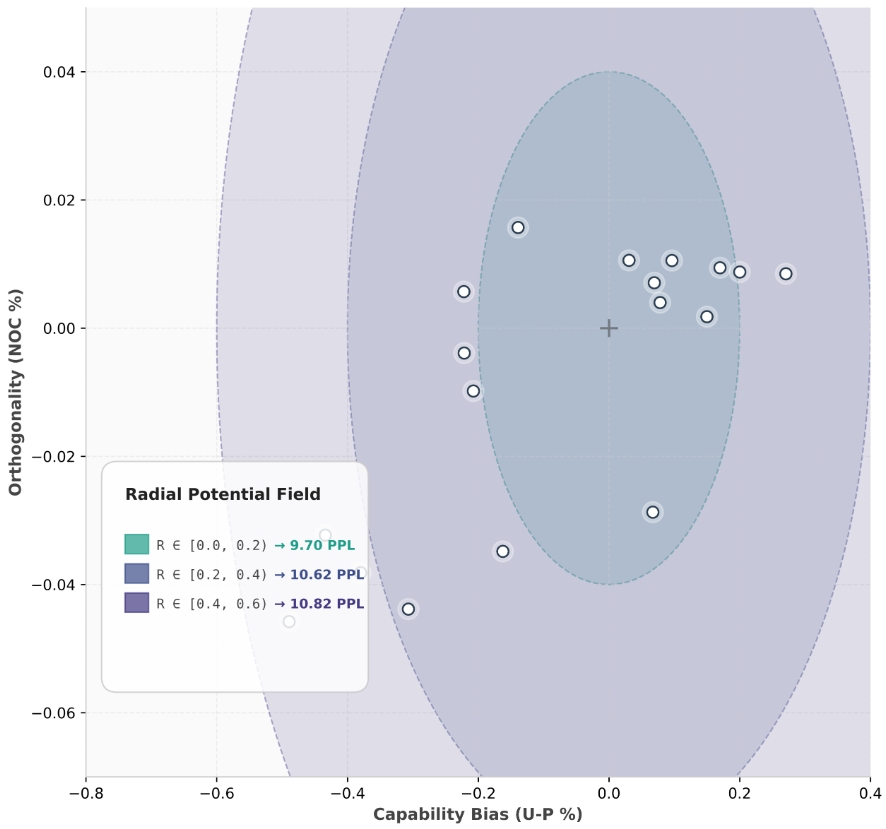}}
    \caption{Distribution of Capability Bias and Orthogonality in Radial Potential Field}
    \label{tu4}
  \vspace{-1em}
\end{figure}
\section{Growth Potential via the Radius $R$}
\label{fuluD}
We define a feature vector $\mathbf{x}=[x_1,x_2,x_3]^{\top}$, where $x_1$, $x_2$, and $x_3$ denote $U_P\%$, NOC$\%$, and performance, respectively. To build a prior indicator, we focus on the subspace $\{U_P,\mathrm{NOC}\}$ that contributes most to structural alignment.

In this subspace, the Euclidean distance to the origin represents the total perturbation energy away from the ideal aligned state at the origin. Using the $\ell_2$ norm,
\begin{equation}
\|\mathbf{x}_{\mathrm{sub}}\|_2=\sqrt{\sum_i \omega_i \hat{x}_i^2},
\end{equation}
where $\hat{x}_i$ are the normalized features.
Without additional prior knowledge, we assume equal contributions from directional shift and structural perturbation. This is consistent with the near-equal loadings observed on the first principal component, so we set $\omega_i=1$ and define
\begin{equation}
R=\sqrt{(U_P\%)^2+(\mathrm{NOC}\%)^2}.
\end{equation}
In Figure \ref{tu4}, the radius $R$ induces level sets on the evolution manifold. Empirically, the expected perplexity satisfies a monotonic relation with $R$:
\begin{equation}
\frac{\partial \mathbb{E}[\mathrm{PPL}]}{\partial R}>0.
\end{equation}
As shown in the Figure \ref{tu4}, Nexusformer follows a centripetal evolution pattern driven by orthogonality constraints. During evolution, points stay close to the orthogonal balance axis and move toward the low-PPL rings near the center. Points farther from the center correspond to higher-PPL regions and weaker quality.
We also validate this trend on downstream performance.
High-efficiency samples concentrate in the core energy region, with a mean radial potential gain of $+5.59\%$.
As the radius increases, both directional shift and structural perturbation grow, and the mean performance decreases in a stepwise manner.

Overall, in the Nexusformer evolution system, the radius $R$ is negatively correlated with growth quality. Smaller $R$ indicates higher-quality growth. By enforcing a centripetal movement pattern, Nexusformer compresses the parameter manifold into a high-energy core region, enabling strong gains at each evolution step.

\section{Computational Cost Analysis and FLOPs Comparison}
\label{fuluE}
\subsection{FLOPs Measurement Methodology}
We conduct a comprehensive measurement of Floating-point Operations (FLOPs) using the CalFlops library. By leveraging a hook mechanism, CalFlops captures the computational workload of all operators during the model's forward pass. This approach enables an accurate accounting of the total computational cost, encompassing linear layers, attention mechanisms, normalization layers, and activation functions.
\subsection{Comprehensive FLOPs Comparison Results}
\begin{figure}[!ht]
   \centering
   \includegraphics[width=\columnwidth]{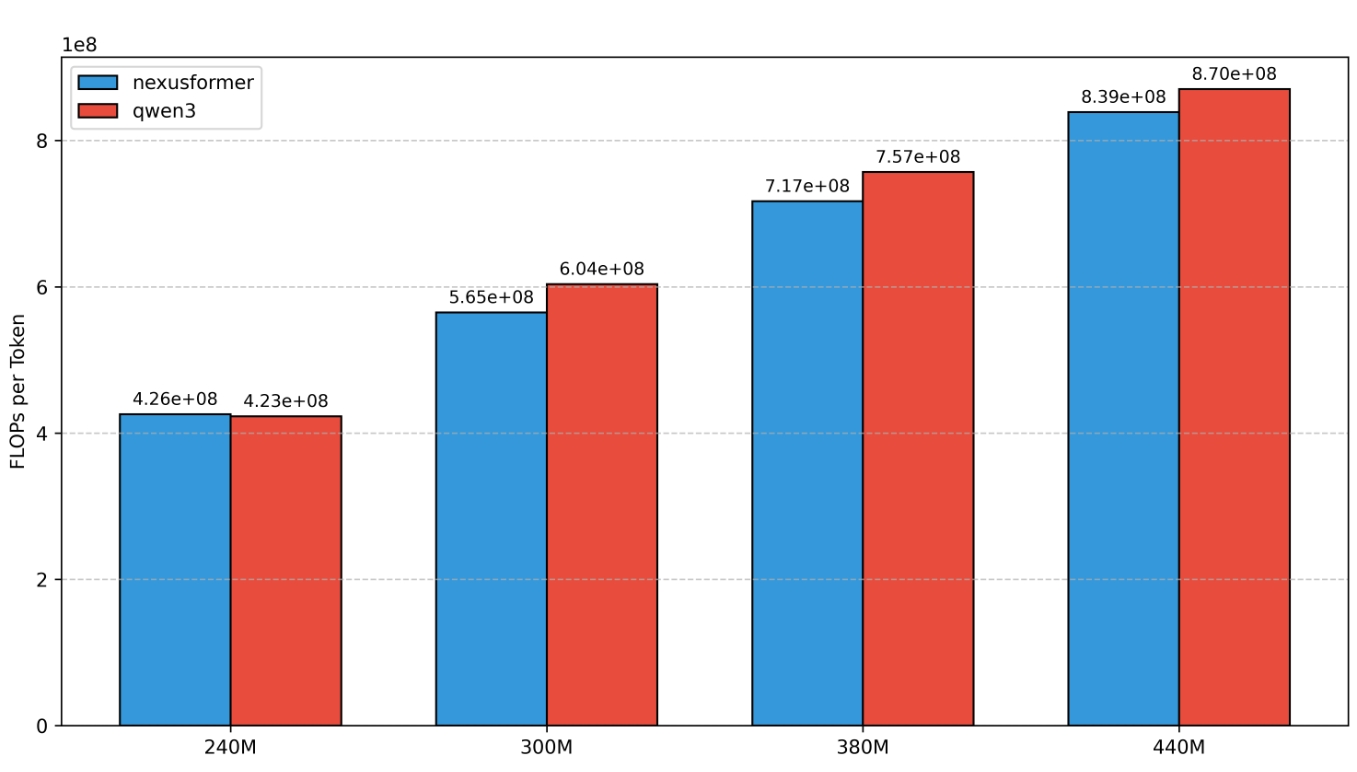}
   \caption{Comparison of Per-Token FLOPs between Nexusformer and Qwen Series across Different Parameter Scales.}
   \label{fig:flops-token}
\end{figure}

Per-Token FLOPs Comparison.Computational costs of Nexusformer and Qwen baseline models across various parameter scales. Although Nexusformer employs a three-stage mapping design (D→M→A→D), its total computational cost is lower than that of the standard single-step mapping (D→D) architecture across most scales.

Precise measurements using CalFlops demonstrate that Nexusformer exhibits superior computational efficiency. At the 240M scale, its FLOPs ($4.26 \times 10^8$) are nearly identical to those of Qwen3; however, as the model scales, the advantages of Nexusformer become more pronounced, achieving a significant 6.5\% reduction in computational workload at the 300M scale. Although this advantage slightly converges to 3.6\% at the 440M scale, Nexusformer consistently maintains lower computational costs than the standard single-step mapping architecture across all medium-to-large configurations. This confirms that its architectural design achieves a robust balance between enhanced representational capacity and computational scalability.

The experiments reveal a pivotal, counter-intuitive finding: despite employing a seemingly more complex three-stage mapping design (D→M→A→D), Nexusformer’s total FLOPs are actually lower than those of the standard single-step mapping (D→D) Transformer at most scales. This efficiency stems from a strategic redistribution of the computational load: by significantly reducing the hidden dimension $D$, Nexusformer drastically compresses the costs of quadratic complexity terms in the attention mechanism—such as $QK^T$ and $AV$ aggregation—which are proportional to $d_{head}$. By trading a higher frequency of "small-scale" linear transformations for substantial savings in high-load computational components, the architecture achieves an optimal equilibrium, delivering superior model performance with lower compute consumption.

\subsection{Trade-off Analysis: Computational Efficiency vs. Perplexity}
While FLOPs serve as a critical metric for measuring computational cost, the true value of a model lies in its modeling performance. This section provides a comprehensive evaluation of Nexusformer’s computational efficiency by integrating FLOPs measurements with Perplexity (PPL) metrics. The integrated comparison results are presented in Table~\ref{tab:flops-ppl}.

\begin{table}[!ht]
\centering
\small
\setlength{\tabcolsep}{6pt}
\renewcommand{\arraystretch}{1.15}
\caption{Comparative analysis of computational efficiency, training cost, and modeling performance between Nexusformer and Qwen series.}
\label{tab:flops-ppl}
\begin{tabular}{l c c c c c}
\hline
Model & Size & FLOPs & Train time & PPL & PPL/FLOPs \\
\hline
nexusformer & 240M & $4.256\times10^8$ & 47    & 10.8157 & $2.54\times10^{-8}$ \\
qwen3         & 240M & $4.223\times10^8$ & 46.1  & 18.1741 & $4.30\times10^{-8}$ \\
nexusformer & 300M & $5.648\times10^8$ & 79.79 & 9.4216  & $1.67\times10^{-8}$ \\
qwen3           & 300M & $6.036\times10^8$ & 87.75 & 13.7495 & $2.28\times10^{-8}$ \\
nexusformer & 380M & $7.171\times10^8$ & 96.3  & 8.3981  & $1.17\times10^{-8}$ \\
qwen3        & 380M & $7.570\times10^8$ & 117   & 12.8842 & $1.70\times10^{-8}$ \\
nexusformer & 440M & $8.390\times10^8$ & 110.36& 9.1706  & $1.09\times10^{-8}$ \\
qwen3         & 440M & $8.704\times10^8$ & 129   & 12.2500 & $1.41\times10^{-8}$ \\
\hline
\end{tabular}
\end{table}
The PPL/FLOPs efficiency ratio serves as a core metric for evaluating a model's computational efficiency, defined as the perplexity achieved per unit of computational cost. A lower ratio signifies higher efficiency. Experimental results demonstrate that Nexusformer exhibits a significant efficiency advantage across all four scales:
\begin{itemize}
    \item 240M Scale: 2.54 vs. 4.30 (40.9\% improvement)
    \item 300M Scale: 1.67 vs. 2.28 (26.8\% improvement)
    \item 380M Scale: 1.17 vs. 1.70 (31.2\% improvement)
    \item 440M Scale: 1.09 vs. 1.41 (22.7\% improvement)
\end{itemize}
This consistent advantage across scales underscores the robustness of the Nexusformer architecture. Notably, the evolution of the efficiency ratio shows that Nexusformer's ratio decreases from 2.54 to 1.09, while the Qwen series decreases from 4.30 to 1.41. The faster rate of decline in Nexusformer indicates a characteristic of increasing marginal returns—the computational efficiency advantage expands as the model scale grows. At the 440M scale, Nexusformer’s efficiency ratio approaches 1.0, implying that each unit of FLOPs yields approximately one unit of perplexity, achieving a near-optimal "compute-performance" equilibrium.

Nexusformer demonstrates a distinct architectural advantage in breaking through the linear computation bottlenecks of traditional models. While conventional Transformers often face a "diminishing marginal utility" in perplexity when scaling up, the rapid decline of Nexusformer’s PPL/FLOPs ratio reflects its exceptional scalability. By maintaining linear computational growth while capturing long-range dependencies more effectively, it surpasses the performance ceilings of similar models under equivalent FLOPs constraints.

The relationship between training time and PPL further reveals that Nexusformer achieves convergence effects far exceeding conventional frameworks within similar or even shorter durations. By optimizing linear layer computation logic, Nexusformer successfully addresses the issue of computational waste during model scaling. The experimental results clearly confirm that Nexusformer can achieve language modeling precision (PPL) that far surpasses traditional mainstream architectures at a lower per-token computational cost (FLOPs/token), effectively realizing a breakthrough in computational efficiency.

\section{Theoretical Analysis of Nonlinear Expansion and Function Approximation}
\label{fuluG}
\subsection{Rank Bottleneck of Standard Linear Projections}
In standard Transformers, attention forms Query, Key, and Value representations through linear projections.
Given an input sequence $X\in\mathbb{R}^{N\times D}$, the projections are
\begin{equation}
Q = XW_Q,\quad K = XW_K,\quad V = XW_V,
\end{equation}
where $W_Q,W_K,W_V\in\mathbb{R}^{D\times D}$ are learnable linear maps.
This design induces an intrinsic representational bottleneck: regardless of optimization, the output is constrained by the column space of the projection.
For any linear map $W\in\mathbb{R}^{D\times D}$, the output rank satisfies
\begin{equation}
\mathrm{rank}(XW)\le \min\big(\mathrm{rank}(X),\,\mathrm{rank}(W)\big).
\end{equation}
Consequently, when a semantic transformation $\mathcal{F}:\mathcal{X}\rightarrow \mathbb{R}^{D}$ requires feature relationships whose effective dimension exceeds $D$, linear projection inevitably discards information.
Moreover, linear maps preserve subspace structure and therefore cannot effectively disentangle nonlinearly separable semantic patterns.
This limitation is most pronounced for higher-order semantic interactions, where complex token dependencies cannot be captured by low-dimensional linear combinations.
We refer to this limitation as the \emph{linear bottleneck}, which upper-bounds the expressivity of attention under a fixed hidden size $D$.

\subsection{Depth Separation: Why Dual Nonlinearity Matters}
To overcome the linear bottleneck, we introduce a deep nonlinear structure.
Depth separation results in modern deep learning theory provide motivation for this design.
Telgarsky (2016) shows that for target functions with compositional structure, deeper networks can achieve the same approximation accuracy with exponentially fewer parameters.
Consider a target semantic transform
\begin{equation}
f^*:\mathbb{R}^{D}\rightarrow \mathbb{R}^{D}
\end{equation}
with a two-stage composition $f^* = g_2\circ g_1$, where
\begin{equation}
g_1:\mathbb{R}^{D}\rightarrow \mathbb{R}^{M},\qquad
g_2:\mathbb{R}^{M}\rightarrow \mathbb{R}^{D}
\end{equation}
correspond to feature extraction and feature composition, respectively.

\paragraph{Compositional efficiency.}
If the target function admits a two-stage compositional structure, then a dual-nonlinear network
\begin{equation}
\mathcal{N}(x)=\sigma\!\big(\sigma(xW_M)W_A\big)W_D
\end{equation}
can achieve approximation error $\epsilon$ with $O(DM+MA+AD)$ parameters.
In contrast, a single-hidden-layer nonlinear network
\begin{equation}
\hat{\mathcal{N}}(x)=\sigma(xW')W''
\end{equation}
requires $O(D^2A/\epsilon)$ parameters to reach the same accuracy.

This result indicates that when semantic transformations are hierarchical---for example, first extracting local patterns and then modeling global interactions among them---dual nonlinearity can approximate the target function with higher parameter efficiency.
In the Nexus-Rank layer, the first activation $\sigma(xW_M)$ disentangles primitive patterns in the intermediate space $\mathbb{R}^{M}$, while the second activation $\sigma((\cdot)W_A)$ models higher-order interactions in the expanded space $\mathbb{R}^{A}$.
This staged modeling strategy aligns with depth separation predictions and motivates the dual-nonlinear design.

\subsection{Difference from Parameter-efficient Fine-tuning}
We further contrast Nexus-Rank with parameter-efficient fine-tuning methods such as LoRA.
LoRA adapts a model by adding a low-rank perturbation $\Delta W = BA$, where $B\in\mathbb{R}^{D\times r}$ and $A\in\mathbb{R}^{r\times D}$ with $r\ll D$.
Its updated projection remains linear:
\begin{equation}
Q_{\mathrm{LoRA}} = X(W_Q+BA)=XW_Q+XBA.
\end{equation}
Therefore, the added capacity is still confined to a linear subspace and cannot escape the rank constraint in Section D.1.
In contrast, Nexus-Rank introduces nonlinear activations in the intermediate stages and maps features onto a nonlinear manifold:
\begin{equation}
Q_{\mathrm{Nexus}}=\sigma\!\big(\sigma(XW_M)W_A\big)W_D.
\end{equation}
From a function-class viewpoint, LoRA remains a linear combination of basis functions, whereas Nexus-Rank forms a deep composition of nonlinear basis functions.
By universal approximation arguments, the latter strictly contains the former and is thus more expressive in principle.

\subsection{Ablation: Necessity of Dual-axis Expansion}
To validate the architectural choices in Nexusformer and their role in stable expansion, we conduct ablations focusing on axis selection during growth (single-axis expansion versus dual-axis synchronous expansion).
All ablations are pretrained on FineWeb under comparable parameter budgets, and perplexity (PPL) is used as the primary metric.
\begin{table}[!ht]
\centering
\small
\setlength{\tabcolsep}{7pt}
\renewcommand{\arraystretch}{1.15}
\caption{Ablation of expansion axes and correlation order (nexusformer-280M).}
\label{tab:axis-ablation}
\begin{tabular}{l c c c c c}
\hline
Model & Axis & Correlation & $M$ & $A$ & PPL \\
\hline
nexusformer-280M & $M$   & $M>A>D$ & 1180 & 960  & 10.89386732 \\
nexusformer-280M & $A$   & $A>M>D$ & 780  & 1360 & 10.89604631 \\
nexusformer-280M & $M{+}A$ & $M>A>D$ & 1130 & 1010 & 10.89386732 \\
nexusformer-280M & $M{+}A$ & $A>M>D$ & 830  & 1310 & 10.89016404 \\
\hline
\end{tabular}
\end{table}

Table~\ref{tab:axis-ablation} shows that performance is jointly shaped by the expansion axis and the correlation order.
First, joint expansion over $M{+}A$ provides better global optimization than single-axis expansion.
Under similar parameter budgets, expanding only $M$ or only $A$ tends to hit a performance ceiling, while $M{+}A$ releases capacity along both axes and offers more flexible representations.
Second, the correlation order affects how capacity is allocated across axes.
With the order $A>M>D$, more capacity is effectively directed to the $A$ axis; combined with joint expansion, this yields the best result, reducing PPL to 10.89.
Overall, the strongest performance arises from the coupling between joint $M{+}A$ expansion and an $A$-guided correlation order, which improves cross-axis information transfer and outperforms symmetric or single-axis growth.

\section{Analysis of empirical phenomena and hyperparameter consistency}
\label{fuluG}
\subsection{Universality of validation-loss rebound under 240M--440M scaling}
Across the 240M--440M expansion experiments, all models exhibit a rebound in validation loss near the end of pretraining.
This can be quantitatively explained by the test-loss shift logic in \cite{fu1}.
In the sample-limited regime with a fixed dataset size $D$, the generalization gap between training loss and test loss follows a power-law increase with model size $N$.
As $N$ grows rapidly, the gap can increase faster than the short-term reduction in training loss, causing the test loss to pass its minimum and show a U-shaped rebound.
This reflects a common bottleneck for high-capacity architectures under limited data density, where additional capacity shifts learning from generalizable patterns toward memorizing sample noise \cite{fu1}.

\subsection{Plateaued scaling from 170M--640M and learning-rate alignment}
The non-linear score scaling observed from 170M to 640M is primarily constrained by the data-limited term in
$L(N,D)=\big[(N_c/N)^{\alpha_N/\alpha_D}+D_c/D\big]^{\alpha_D}$ \cite{fu1}.
Notably, we use a fixed learning rate (LR) for all model sizes.
According to Yang et al.~\cite{fu2}, under standard parameterization the optimal LR shifts with model width.
While this choice is not compute-optimal from a scaling-law perspective, we intentionally keep hyperparameters consistent to avoid gains from per-scale tuning (e.g., $\mu$P transfer).
Under this controlled setting, the observed plateau reflects the true performance floor imposed by the current data scale $D$.
This design allows us to evaluate Nexusformer's architectural improvements under limited information entropy, rather than attributing gains to scaling or hyperparameter optimization.


\end{document}